\documentclass{article} 
\usepackage{iclr2027_preprint,times,amssymb}

\usepackage{amsmath,amsfonts,bm}

\def\eqref#1{equation~\ref{#1}}

\def\1{\bm{1}}

\DeclareMathAlphabet{\mathsfit}{\encodingdefault}{\sfdefault}{m}{sl}
\SetMathAlphabet{\mathsfit}{bold}{\encodingdefault}{\sfdefault}{bx}{n}

\usepackage{hyperref}
\usepackage{url}
\usepackage{booktabs}
\usepackage{multirow}
\usepackage{graphicx}
\usepackage{array}
\usepackage[table]{xcolor}
\usepackage{wrapfig}
\usepackage{tabularx}
\usepackage{MnSymbol}
\usepackage{float}
\usepackage{enumitem}
\usepackage{pifont}
\usepackage{microtype}
\newcommand{\cmark}{\ding{51}}
\newcommand{\xmark}{\ding{55}}
\definecolor{headergray}{HTML}{E4EAF1}
\definecolor{groupgray}{HTML}{E4EAF1}
\definecolor{subgroupgray}{HTML}{F3F5F8}
\definecolor{summaryblue}{HTML}{EDF3FA}

\definecolor{headergray}{gray}{0.92}
\definecolor{oursblue}{RGB}{225,235,250}

\usepackage{tikz}
\usepackage{xcolor}

\definecolor{SIbg}{HTML}{DCEAF7}
\definecolor{SIfg}{HTML}{356A92}

\definecolor{SXbg}{HTML}{F7EDC7}
\definecolor{SXfg}{HTML}{8A6A1F}

\definecolor{DIbg}{HTML}{F6DADA}
\definecolor{DIfg}{HTML}{9A4747}

\definecolor{DXbg}{HTML}{DDEEDC}
\definecolor{DXfg}{HTML}{4F7A50}

\definecolor{bestcolor}{HTML}{edadae}
\colorlet{secondcolor}{bestcolor!50!white}
\definecolor{headerbg}{HTML}{D9E2EF}     
\definecolor{groupgray}{HTML}{E6E9EF}    
\definecolor{subgroupgray}{HTML}{F3F4F7} 
\definecolor{summaryblue}{HTML}{EDF2FA}  

\newcommand{\best}[1]{\cellcolor{bestcolor}\textbf{#1}}
\newcommand{\second}[1]{\cellcolor{secondcolor}\underline{#1}}
\newcommand{\groupsep}{\midrule[0.6pt]}
\newcommand{\subsep}{\arrayrulecolor{black!30}\midrule[0.3pt]\arrayrulecolor{black}}

\newcommand{\methodtag}[3]{%
  \tikz[baseline=(tag.base)]{
    \node[
      anchor=base,
      rounded corners=1.5pt,
      inner xsep=2.5pt,
      inner ysep=0.6pt,
      fill=#1,
      text=#2,
      font=\scriptsize\sffamily\bfseries
    ] (tag) {#3};
  }%
}

\newcommand{\SItag}{\methodtag{SIbg}{SIfg}{SI}}
\newcommand{\SXtag}{\methodtag{SXbg}{SXfg}{SX}}
\newcommand{\DItag}{\methodtag{DIbg}{DIfg}{DI}}
\newcommand{\DXtag}{\methodtag{DXbg}{DXfg}{DX}}

\begin{document}

\textls[-25]{\title{GraphMAS: A Systematic Benchmark of Multi-Agent Coordination for Graph Learning}}


\author{
Jiayi Yang\textsuperscript{1},
Yifang Chen\textsuperscript{1},
Yuanfu Sun\textsuperscript{2},
Xinyan Ge\textsuperscript{3},
Qiaoyu Tan\textsuperscript{1\thanks{Corresponding author.}}
\\
\textsuperscript{1}New York University Shanghai
\quad
\textsuperscript{2}New York University
\quad
\textsuperscript{3}Northwestern University
\\
\texttt{\{jy4656,qiaoyu.tan\}@nyu.edu}
}
%

\newcommand{\fix}{\marginpar{FIX}}
\newcommand{\new}{\marginpar{NEW}}

\maketitle

\begin{abstract}
Large language model (LLM)-based multi-agent systems coordinate specialized reasoning through aggregation, interaction, and adaptive control, yet their potential for graph learning remains largely unexplored. Graph learning is a natural setting for such systems because useful evidence may arise from heterogeneous local, long-range, global structural, and semantic perspectives whose relevance varies across instances. Existing LLM-based graph learning approaches primarily rely on single-agent reasoning, while multi-agent coordination has been studied mainly in general reasoning settings. Consequently, it remains unclear whether multiple specialized agents can improve graph learning and how different coordination strategies should be designed and evaluated. To address this gap, we introduce \textsc{GraphMAS}, a systematic benchmark of multi-agent coordination for graph learning. \textsc{GraphMAS} builds a shared pool of graph reasoning specialists and organizes coordination along two dimensions, inter-agent interaction and runtime adaptivity, yielding four paradigms and seven representative coordination methods. Under a unified protocol, we evaluate these methods across seven text-attributed graphs, three domains, and two representative graph learning tasks: node classification and link prediction. Our study shows that heterogeneous graph perspectives are strongly complementary, and that coordinating specialists improves over both individual specialists and single-agent graph reasoning methods, with gains that arise from decomposing reasoning across specialists rather than from broader evidence access alone. However, richer inter-agent interaction does not reliably help, whereas instance-adaptive specialist selection yields the strongest accuracy–efficiency trade-off. We further show that the coordination policy can be learned over a fixed specialist pool and transfers to held-out graphs. \textsc{GraphMAS} therefore provides a controlled evaluation framework and empirical principles for understanding when and how multi-agent coordination benefits graph learning.

\end{abstract}

\section{Introduction}
\label{sec:introduction}
LLM-based multi-agent systems coordinate multiple
reasoning processes through role specialization, independent judgments,
aggregation, and exchanges of intermediate outputs. Such systems have shown
promising results in domains including software development and mathematical
reasoning~\citep{hong2024metagpt,du2024debate}, motivating a growing range of
coordination strategies with different communication structures and agent
participation rules~\citep{tran2025collaboration}. These strategies determine
which agents contribute, what information they share, and when reasoning
terminates. While multi-agent coordination has been increasingly studied in
general reasoning settings, its potential for \emph{graph learning} remains
largely unexplored.


Graph learning provides a natural setting for studying multi-agent
coordination because useful evidence can arise from heterogeneous graph
perspectives whose relevance varies across instances. Predictions on
text-attributed graphs may depend on local neighborhoods, longer-range
structural relationships, graph-wide relevance, or node semantics.
Classical graph learning has long recognized that different receptive fields
and structural contexts can be useful for different nodes, for example by
adapting neighborhood ranges~\citep{xu2018representation} or combining local
message passing with global attention~\citep{rampasek2022recipe}. More
recently, LLM-based graph learning methods have incorporated graph structure
through instruction tuning, structured prompting, and active graph
exploration~\citep{chen2024llaga,sun2025graphicl,liu2026graphsearch,sun2026agentgl}.
However, these approaches primarily rely on a \emph{single reasoning agent} to
integrate heterogeneous structural and semantic evidence within one
trajectory. When different graph perspectives provide complementary signals,
a single reasoning process may not exploit them equally well.

\begin{wrapfigure}{r}{0.46\textwidth}
\vspace{-1.2em}
    \label{fig:oracle}
    \centering
    \setlength{\abovecaptionskip}{3pt}
    \setlength{\belowcaptionskip}{0pt}
    \includegraphics[width=\linewidth]{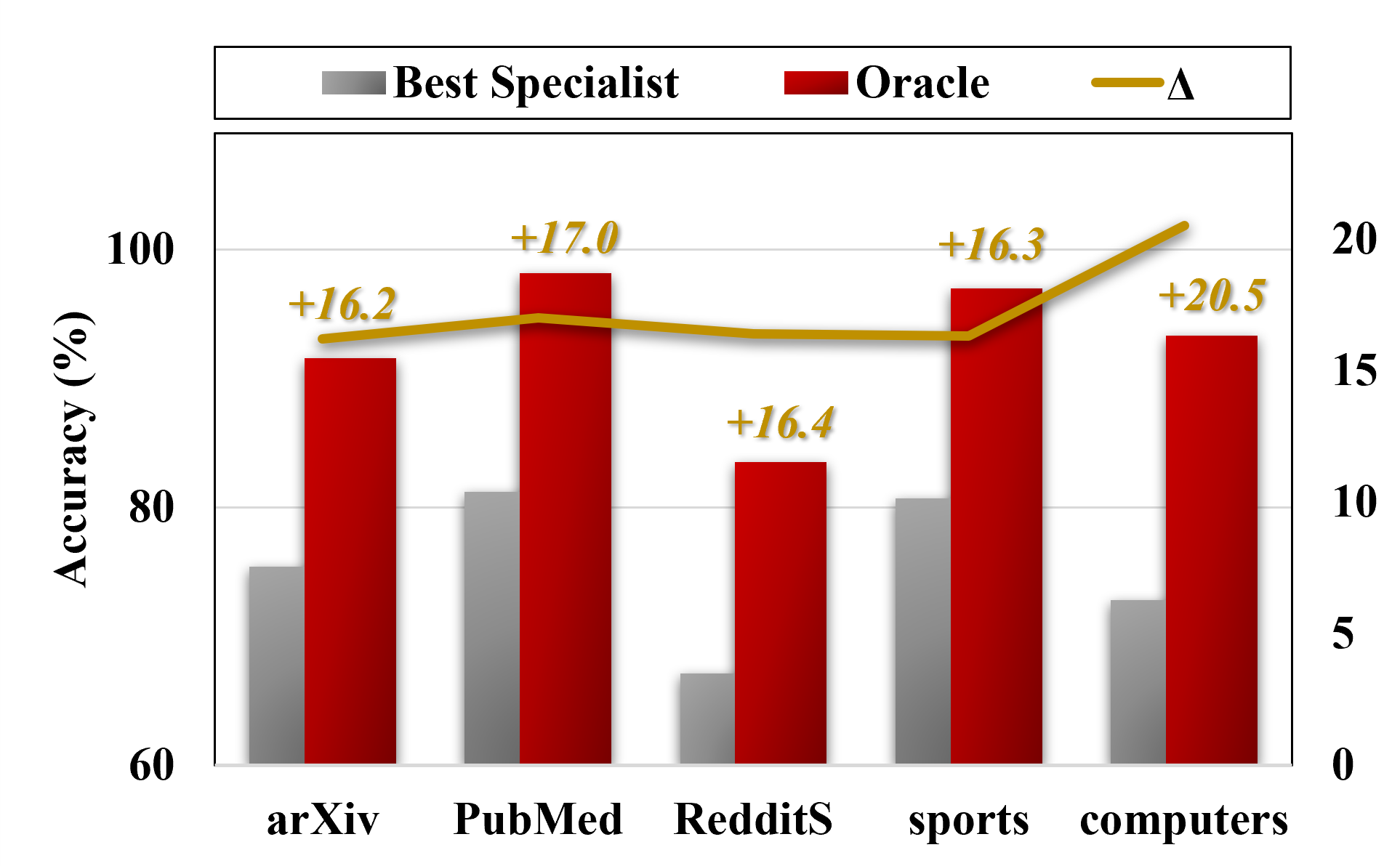}
    \caption{Performance gap between the best specialist and oracle on five LP tasks. The oracle is correct if any specialist is correct; the best specialist is selected per dataset. Exact results are reported in Appendix~\ref{app:oracle}.}
\vspace{-0.8em}
\end{wrapfigure}

This observation motivates a different formulation: assigning distinct graph
perspectives to specialized reasoning agents and coordinating their judgments
for a shared graph learning task. Specialization, however, creates a new
coordination problem. An evidence perspective that is informative for one
instance may be unhelpful for another, and specialists may therefore exhibit
both complementary strengths and conflicting predictions. As illustrated by the oracle gap in Figure~\ref{fig:oracle}, specialist
success varies across instances, leaving considerable \emph{coordination headroom}: instances missed by the best individual
specialist can often be resolved by another graph perspective. The central
question is therefore not simply whether more agents improve graph learning,
but \emph{when complementary specialists should participate and how their
reasoning should be coordinated}.


Despite this opportunity, systematically understanding multi-agent
coordination for graph learning remains challenging for three reasons. \textbf{1) There is no controlled benchmark for comparing coordination
strategies in graph learning.}
Existing LLM-based graph learning methods differ substantially in how they
access and reason over graph information, while general multi-agent methods
differ in agent participation, communication, and runtime control.
Consequently, directly comparing complete systems can conflate the effects of
the underlying graph reasoning capability with those of the coordination
mechanism. A meaningful comparison therefore requires a shared specialist
pool, consistent graph evidence interfaces, and controlled inference
conditions. \textbf{2) The source of multi-agent gains is unclear.}
Improvement from multiple agents may arise from several factors: access to
different graph evidence, decomposition of reasoning across multiple
trajectories, specialized role conditioning, or the coordination mechanism. Moreover, comparisons with individual specialists alone cannot
establish an advantage over strong single-agent graph learning systems.
Controlled analysis is therefore needed to determine whether multi-agent
reasoning provides benefits beyond broader evidence access and to understand
how graph evidence and specialist roles interact. \textbf{3) It remains unclear when additional coordination is actually
useful.}
Multi-agent systems often increase communication or invoke additional agents
uniformly across instances. Yet the need for coordination is inherently
instance dependent: specialists may already agree on easy cases, while
disagreement may indicate that different graph perspectives support competing
predictions. More interaction may therefore increase computation without
necessarily improving accuracy. Understanding which specialists should participate and whether they should communicate is essential for both effective and efficient multi-agent graph learning.


To address these challenges, we introduce \textbf{GraphMAS}, a systematic
benchmark of multi-agent coordination for graph learning. GraphMAS evaluates
representative coordination approaches over a shared pool of graph reasoning
specialists under consistent experimental conditions. The specialists capture
four complementary graph perspectives---proximal neighborhoods, distal
neighborhoods, graph-wide structural relevance, and semantic affinity---while
sharing the same agentic backbone, retrieval budget, and LLM configuration.
We organize coordination along two dimensions:
\textbf{inter-agent interaction}, which distinguishes independent specialist
reasoning from reasoning conditioned on other specialists' intermediate
outputs; and \textbf{runtime adaptivity}, which distinguishes predefined
coordination procedures from instance-dependent participation, communication,
or stopping decisions. Crossing these dimensions yields four paradigms:
Static Independent, Static Interactive, Dynamic Independent, and Dynamic
Interactive. GraphMAS instantiates seven representative coordination methods
across these paradigms and evaluates them alongside strong LLM reasoning,
agentic graph learning, and individual specialist baselines on node
classification and link prediction.


Our study reveals several consistent empirical patterns. 
First, heterogeneous graph perspectives exhibit substantial instance-dependent complementarity, creating clear headroom for coordination beyond any individual specialist.
Second, controlled ablations show that agent specialization is most effective when graph evidence and reasoning roles are differentiated jointly, and distributing complementary perspectives across specialists provides gains beyond single-agent reasoning over the same evidence sources.
Third, richer inter-agent interaction does not consistently
improve performance or efficiency; instead, instance-adaptive specialist selection achieves a favorable accuracy--efficiency trade-off, with coordination methods differentiating most strongly under specialist disagreement.
Finally, we show that the coordination policy itself can be learned while keeping the underlying graph specialists fixed, and that learned coordination can transfer to held-out graphs.


Our main \textbf{contributions} are summarized as follows:

\vspace{-0.5em}

\begin{itemize}[leftmargin=*]
        \item \textbf{A systematic benchmark for multi-agent coordination in graph learning.}
    We introduce GraphMAS, which studies how a shared pool of graph reasoning specialists can be coordinated for graph learning. We characterize
    coordination along inter-agent interaction and runtime
    adaptivity, yielding four paradigms instantiated by seven representative
    coordination methods.

    \item \textbf{A unified and comprehensive evaluation protocol.}
    GraphMAS evaluates multi-agent coordination methods across seven text-attributed graphs spanning three domains and two representative graph learning tasks. The benchmark measures
    predictive performance and compares against strong
    non-agentic, agentic, and individual-specialist baselines.

    \item \textbf{Controlled diagnostic analyses.}
    We isolate the contributions of evidence allocation, role specialization, and reasoning decomposition through controlled ablations, and analyze coordination behavior through oracle gaps, agreement-stratified evaluation, and cost accounting.

    \item \textbf{Extension to learned coordination.}
    We demonstrate that coordination can be optimized while keeping graph specialists fixed, establishing learned routing as a strong extension of the benchmark and showing that coordination behavior can generalize on unseen graphs.
\end{itemize}




\section{Related Work}
\label{sec:related_work}

\paragraph{Multi-Agent LLM Systems.}
Multi-agent LLM systems coordinate agents through task decomposition,
role specialization, and intermediate exchanges~\citep{tran2025collaboration}.
Such coordination has improved software-generation quality in
MetaGPT~\citep{hong2024metagpt} and reasoning accuracy and factuality
in Multi-Agent Debate~\citep{du2024debate}.
Despite these shared benefits, existing systems adopt different mechanisms
for information exchange and execution control.
Some aggregate independently generated
responses~\citep{jiang2023llmblender,li2024more}, while others incorporate
one agent's output into another's reasoning.
Participation and communication may follow predefined protocols or
depend on model-generated control decisions.
For example, DyLAN~\citep{liu2024dylan} selects agents through response
rankings, while Graph-of-Agents~\citep{yun2026goa} uses response evaluations
to select relevant agents and construct communication graphs.
Inter-agent interaction and runtime adaptivity are thus distinct.
Fixed workflows can support intermediate exchanges, while adaptive
selection can preserve independent reasoning.

\paragraph{LLM-based Graph Learning.}
LLM-based graph learning requires integrating attribute semantics with
relational structure~\citep{jin2024large}.
GraphGPT~\citep{tang2024graphgpt}, LLaGA~\citep{chen2024llaga}, and
GraphICL~\citep{sun2025graphicl} address this challenge through graph
instruction tuning, structure-aware projection, and structured prompting,
respectively.
Agentic methods further acquire graph evidence during inference.
Graph-CoT~\citep{jin2024graphcot} interleaves reasoning with graph
interaction, while GraphSearch~\citep{liu2026graphsearch} and
AgentGL~\citep{sun2026agentgl} use graph-aware retrieval and learned
navigation, respectively.
More recently, GLM~\citep{huan2025scaling} and
MAAGL~\citep{qu2026maagl} introduced specialized-agent designs for graph
question answering and node classification, respectively.
Although these studies demonstrate the potential of multi-agent graph
reasoning, their focus on specific architectures and task settings leaves
systematic comparisons across coordination paradigms and graph learning
tasks limited.

\section{GraphMAS}

\vspace{-0.5em}

GraphMAS provides a systematic framework for studying multi-agent coordination
in graph learning. As illustrated in
Figure~\ref{fig:graphmas_overview}, it builds on a shared pool of graph
reasoning specialists with heterogeneous graph perspectives, detailed in Section~\ref{sec:specialists}. Section~\ref{sec:design-space} then
characterizes the coordination space along two dimensions, and
Section~\ref{sec:methods} instantiates representative methods within this
space. 

\vspace{-0.5em}

\subsection{Preliminaries}
\label{sec:prelim}
\vspace{-0.5em}

We consider a text-attributed graph
$\mathcal{G}=(\mathcal{V},\mathcal{E},\mathcal{T})$, where each node
$v\in\mathcal{V}$ carries a textual attribute $t_v\in\mathcal{T}$, and study
two representative tasks. In node classification (NC), the goal is to
predict the label $y_v\in\mathcal{Y}$ of a target node $v$. In link
prediction (LP), the goal is to predict whether an edge exists between a node
pair $(u,v)$.
We write $q$ for a generic prediction instance, a target node for NC or a node
pair for LP, and $y_q$ for its ground-truth label, which is never observable
during inference. 

\vspace{-0.5em}
\subsection{Graph Specialists}
\label{sec:specialists}
\vspace{-0.5em}

Graph prediction may require local, distant, or semantically related evidence, with the most useful perspective varying by instance. Earlier LLM-based graph methods use fixed-hop subgraphs or predefined neighborhood templates~\citep{10.1145/3589335.3651476,chen2024llaga}, while agentic methods let a single agent search across structural and semantic perspectives~\citep{liu2026graphsearch,sun2026agentgl}. Within one reasoning trajectory, however, complementary perspectives may receive uneven attention, making their contributions difficult to distinguish. GraphMAS assigns these perspectives to specialized agents and studies how to coordinate their predictions.

\vspace{-1em}

\paragraph{Shared agentic backbone.}
 All specialists use the same retrieval
pipeline, search budget, decoding configuration, and a frozen LLM backbone consistent with prior graph learning works~\citep{liu2026graphsearch}. Given an instance $q$, a specialist alternates between reasoning, issuing graph queries, and inspecting
the returned evidence through a
\texttt{<think>}\,/\,\texttt{<search>}\,/\,\texttt{<information>} interaction
loop until it produces a prediction or reaches its search budget. Each query is
handled by a graph-aware retriever that constructs candidates according to the
specialist's requested evidence scope and ranks them using structural and
semantic relevance signals.
\label{sec:graphmas}
\begin{figure*}[t]
    \centering
    \includegraphics[width=\textwidth]{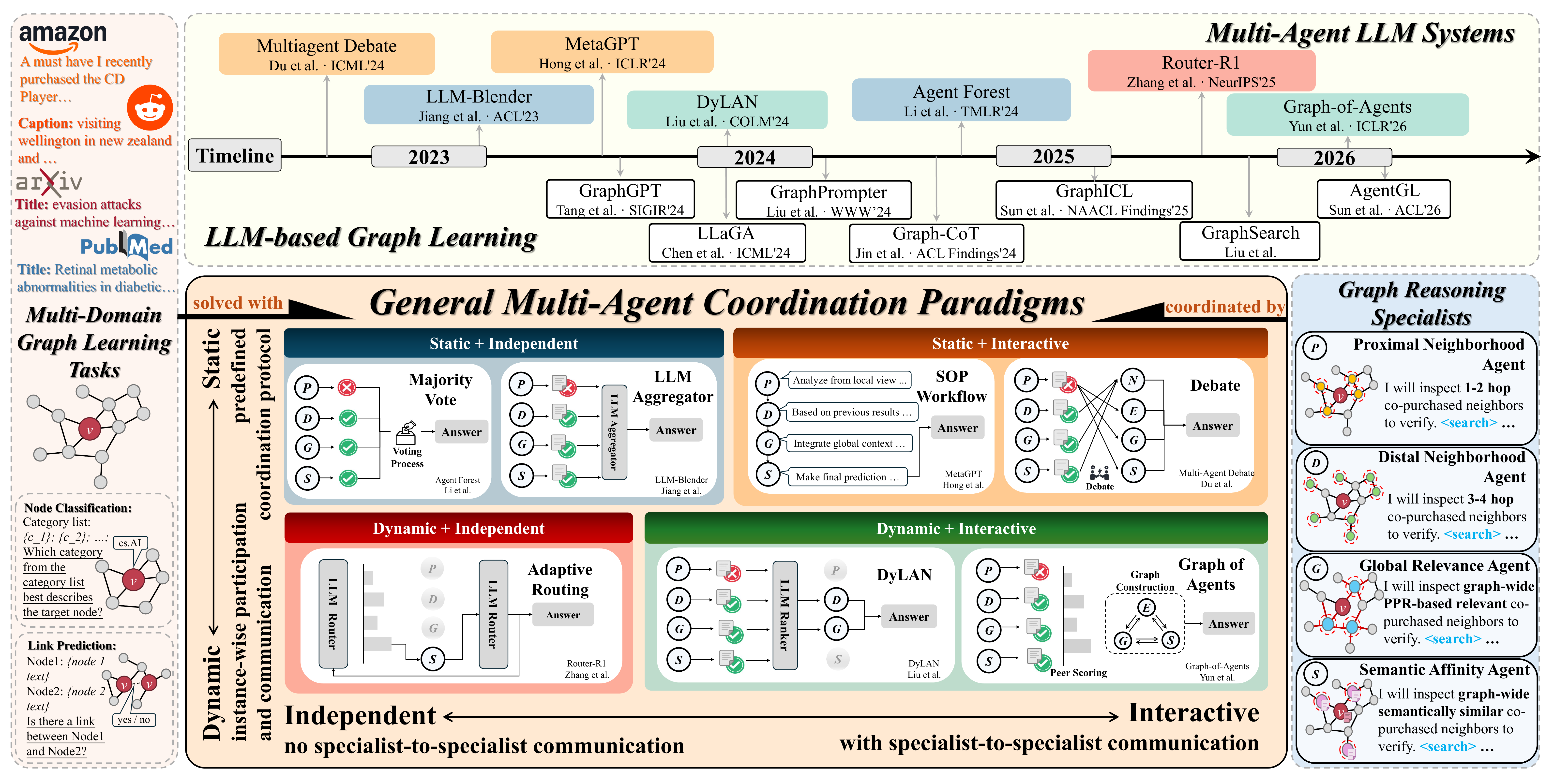}
    \vspace{-1.5em}
    \caption{
Overview of GraphMAS. A shared pool of graph reasoning specialists is
coordinated across representative multi-agent paradigms organized by
\textbf{inter-agent interaction} and \textbf{runtime adaptivity}. The timeline
shows prior LLM-based graph learning and general multi-agent systems.
    }
    \vspace{-1em}
    \label{fig:graphmas_overview}
\end{figure*}

\vspace{-1em}

\paragraph{Specialized roles.}
We select specialists based on \textbf{structural relationships} and
\textbf{semantic similarity}.
Structural exploration separates proximal and distal neighborhoods to
examine local and longer-range context independently, complemented by
graph-wide relevance without a fixed hop constraint.
Semantic exploration retrieves textually related nodes regardless of graph proximity.
We assign these four views to separate specialists, building on the retrieval
modes of prior agentic graph learning~\citep{liu2026graphsearch,sun2026agentgl},
and study how to coordinate their predictions:

\vspace{-0.5em}

\begin{itemize}
    \item \textbf{Proximal Neighborhood Agent} focuses on relational evidence
    close to $q$. Nearby nodes often provide direct contextual signals through homophily, shared interactions, or local
    structural patterns. We instantiate this perspective using 1- and 2-hop
    retrieval.

    \item \textbf{Distal Neighborhood Agent} examines relational evidence
    beyond the immediate neighborhood, capturing dependencies that require a
    broader structural receptive field. We instantiate this perspective using
    3- and 4-hop retrieval excluding the proximal neighbors.

    \item \textbf{Global Relevance Agent} targets structurally relevant nodes
    that need not fall within a fixed hop radius. We operationalize this
    perspective by ranking graph-wide candidates according to Personalized
    PageRank~\citep{page1999pagerank} with respect to the target, thereby prioritizing nodes with high
    structural relevance even when they are topologically distant.

    \item \textbf{Semantic Affinity Agent} prioritizes attribute-level relevance over graph proximity, retrieving graph-wide nodes with semantically similar textual attributes, including informative examples that may be weakly connected or structurally distant.
\end{itemize}

\vspace{-0.5em}

For a node-classification instance $q=v$, each specialist applies its
corresponding scope with respect to the target node $v$. For a
link-prediction instance $q=(u,v)$, the same scope is applied to both
endpoints, and the retrieved evidence is presented jointly to the specialist. Together, these agents enable broad graph exploration through distinct yet complementary structural and semantic perspectives.

\vspace{-0.5em}

\paragraph{Specialist interface.}
To support different coordination mechanisms, all specialists expose a common interface. A single invocation of specialist $A_i$ receives a prediction
instance $q$ and an optional coordination context $z$, and returns

\vspace{-0.5em}
\begin{equation}
    o = A_i(q,z) = (\hat{y}, r),
    \label{eq:agent}
\end{equation}

\vspace{-0.5em}
where $\hat{y}$ is the predicted label and $r$ is a textual response
summarizing the evidence and reasoning supporting the prediction. When $z=\varnothing$, the specialist reasons independently. Otherwise, $z$
contains responses made available by the coordination mechanism. The
specialist may use these responses to reconsider its prediction, while graph evidence remains restricted to its assigned perspective. This interface allows the same specialist pool to be used across heterogeneous coordination structures.

\vspace{-0.5em}

\subsection{The Coordination Design Space}
\label{sec:design-space}
\vspace{-0.5em}

Given a specialist pool $\mathcal{A}=\{A_1,\ldots,A_M\}$, we represent the
execution of a multi-agent system on an instance $q$ as a typed coordination
graph

\vspace{-0.5em}
\begin{equation}
    \mathcal{H}(q)
    =
    \big(
    \mathcal{N}_q,
    \mathcal{E}^{\mathrm{msg}}_q,
    \mathcal{E}^{\mathrm{ctrl}}_q
    \big).
    \label{eq:coord-graph}
\end{equation}

\vspace{-0.5em}

The node set contains both specialist and coordination invocations,
$\mathcal{N}_q=\mathcal{N}^{\mathrm{sp}}_q
\cup\mathcal{N}^{\mathrm{coord}}_q$.
A specialist node $n\in\mathcal{N}^{\mathrm{sp}}_q$ invokes a specialist
$A_{a(n)}$ and produces $o_n=A_{a(n)}(q,z_n)$. A coordination node
$n\in\mathcal{N}^{\mathrm{coord}}_q$ invokes a graph-blind coordination
module $C_{c(n)}$, such as a router, aggregator, or ranker, and
produces $o_n=C_{c(n)}(q,z_n)$. Coordination modules may condition on the
prediction instance, specialist descriptions, and responses made available
to them, but cannot invoke retrieval on graphs.

The two edge types distinguish information flow from execution control.
A message edge $(n'\!\rightarrow n)\in\mathcal{E}^{\mathrm{msg}}_q$ indicates
that $o_{n'}$ is included in the reasoning context $z_n$. A control edge
$(n'\!\rightarrow n)\in\mathcal{E}^{\mathrm{ctrl}}_q$ indicates that the
output of $n'$ affects whether or how $n$ is executed, for example through
routing, selection, pruning, or stopping. The final prediction is obtained by
a fixed readout $\rho$ over the realized execution,
$\hat{y}_q=\rho(\mathcal{H}(q),\{o_n\}_{n\in\mathcal{N}_q})$. Any
LLM-based aggregation or synthesis is represented explicitly as a coordination
node rather than being absorbed into $\rho$.

\vspace{-1em}

\paragraph{Inter-agent interaction.}
A mechanism is \emph{interactive} if the response of one specialist can enter
the reasoning context of another specialist before the latter produces or
revises its prediction. It is \emph{independent} otherwise. Coordination
modules may therefore aggregate, rank, or route specialist responses without
making the specialists themselves interactive.

\vspace{-1em}

\paragraph{Runtime adaptivity.}
A mechanism is \emph{static} when specialist participation and communication
follow a predefined protocol shared across instances, possibly with fixed
deterministic stopping rules. A mechanism is \emph{dynamic} when
model-generated control decisions determine part of the realized execution
graph, such as which specialists are invoked or which communication links are
formed.

Crossing the two dimensions yields four coordination paradigms:
\SItag\ \textbf{Static Independent}, \SXtag\ \textbf{Static Interactive},
\DItag\ \textbf{Dynamic Independent}, and \DXtag\ \textbf{Dynamic Interactive}.

\vspace{-0.5em}

\subsection{Instantiated Coordination Methods}
\label{sec:methods}
\vspace{-0.5em}

We instantiate seven representative coordination methods spanning the four paradigms over the same specialist pool, as illustrated in Figure~\ref{fig:graphmas_overview}. They differ in specialist
participation, inter-specialist communication, runtime adaptation, and
final decision-making. Table~\ref{tab:paradigms} summarizes these structures.

\begin{table}[t]
\vspace{-2em}
\centering
\caption{Coordination methods instantiated in GraphMAS.}
\label{tab:paradigms}
\small
\setlength{\tabcolsep}{3.2pt}
\resizebox{\linewidth}{!}{%
\begin{tabular}{@{}llllll@{}}
\toprule
\textbf{Paradigm} & \textbf{Method} & \textbf{Participation}
& \textbf{Specialist communication} & \textbf{Runtime adaptation} & \textbf{Final decision} \\
\midrule
\multirow{2}{*}{\SItag\ \textbf{Static Independent}}
& \emph{Majority Vote} & all $M$ & none & none & plurality vote \\
& \emph{LLM Aggregation} & all $M$ & none & none & LLM aggregator \\
\midrule
\multirow{2}{*}{\SXtag\ \textbf{Static Interactive}}
& \emph{SOP} & all $M$ in fixed order & sequential & none & final specialist \\
& \emph{Debate} & all $M$ per round & all-to-all & none & plurality vote \\
\midrule
\DItag\ \textbf{Dynamic Independent}
& \emph{Adaptive Routing} & selected sequentially & none
& routing and stopping & router \\
\midrule
\multirow{2}{*}{\DXtag\ \textbf{Dynamic Interactive}}
& \emph{DyLAN} & all $M$, then top-$k$ & all-to-all
& agent pruning & agreement/ranker \\
& \emph{Graph of Agents} & all $M$, then top-$k$
& weighted directed graph & graph construction and pruning
& meta-LLM pooling \\
\bottomrule

\end{tabular}%
}
\vspace{-1em}
\end{table}

\vspace{-0.3em}

\paragraph{Static Independent.}
All specialists participate according to a fixed protocol and produce their
responses without observing the reasoning contexts of one another.

\vspace{-0.3em}

\SItag\ \textbf{Majority Vote} follows the sampling-and-voting paradigm of Agent
Forest~\citep{li2024more}. Each specialist independently produces
$o_i=A_i(q,\varnothing)$, and the final prediction is
$\hat{y}_q=\operatorname{mode}\{\hat{y}_i\}_{i=1}^{M}$.

\vspace{-0.3em}

\SItag\ \textbf{LLM Aggregation} adapts the generative-fusion component of
LLM-Blender~\citep{jiang2023llmblender}. Each specialist independently
produces a response $o_i$, after which an aggregator node conditions on the
original instance and the response set to generate the final prediction
$\hat{y}_q=C_{\mathrm{agg}}\!\left(q,\{o_i\}_{i=1}^{M}\right)$.

\vspace{-0.3em}

\paragraph{Static Interactive.}
All specialists participate under a fixed communication structure, and
subsequent specialist invocations may condition on responses produced
earlier in the execution.

\vspace{-0.3em}

\SXtag\ \textbf{SOP Workflow} is inspired by the standardized operating procedures of
MetaGPT~\citep{hong2024metagpt}. Specialists act in the fixed order
$\pi=(\textsc{Proximal},\textsc{Distal},\textsc{Global},\textsc{Semantic})$.
At step $t$, specialist $\pi_t$ observes all preceding responses and produces
$o_{\pi_t}=A_{\pi_t}(q,\{o_{\pi_s}\}_{s<t})$.
The response of the final specialist is returned as the prediction.

\vspace{-0.3em}

\SXtag\ \textbf{Multi-Agent Debate} adapts the iterative debate protocol of multiagent debate~\citep{du2024debate}. Specialists first respond independently,
$o_i^{(0)}=A_i(q,\varnothing)$. At revision round $t$, each specialist
observes its peers' responses and produces
$o_i^{(t)}=A_i(q,\{o_j^{(t-1)}\}_{j\neq i})$ with the assigned graph
perspective unchanged. Debate terminates upon unanimous agreement or after $R$
revision rounds. If disagreement remains, the final prediction is determined by plurality over the specialists' latest predictions.

\vspace{-0.3em}

\paragraph{Dynamic Independent.}
Specialist participation is determined at runtime, while each invoked
specialist reasons without access to other specialists' responses.

\vspace{-0.3em}

\DItag\ \textbf{Adaptive Routing} is conceptually inspired by the sequential routing formulation of
Router-R1~\citep{zhang2025routerr1}. At step $t$, the router conditions on the
instance, specialist descriptions, and collected responses, and either selects
an unqueried specialist $a_t$ or stops with a final prediction. A selected
specialist is invoked independently as $o_t=A_{a_t}(q,\varnothing)$. If the
budget is exhausted, the same coordination model synthesizes the collected
responses into the final answer.

\vspace{-0.3em}

\paragraph{Dynamic Interactive.}
Specialist participation and communication structure may be determined
during execution, and active specialists can revise their predictions using other specialists' responses.

\vspace{-0.3em}

\DXtag\ \textbf{DyLAN}~\citep{liu2024dylan} adapts temporal feed-forward interaction
and inference-time team reformation to the GraphMAS specialist pool. Specialists
first produce independent responses, $o_i^{(0)}=A_i(q,\varnothing)$. If at
least three agree, inference terminates. Otherwise, a listwise ranker selects
$S=C_{\mathrm{rank}}(q,\{o_i^{(0)}\}_{i=1}^{M})$, with $|S|=k$. The retained
specialists then revise their predictions using the other retained responses,
$o_i^{(1)}=A_i(q,\{o_j^{(0)}\}_{j\in S\setminus\{i\}})$. The system terminates
upon agreement. Otherwise, the ranker selects among the final retained responses.

\vspace{-0.3em}

\DXtag\ \textbf{Graph of Agents}~\citep{yun2026goa} is adapted to the GraphMAS
specialist pool using response-conditioned graph construction and message
passing. Specialists first produce independent responses,
$o_i^{(0)}=A_i(q,\varnothing)$, and score the relevance of one another's
responses. The top-$k$ specialists form a weighted directed
graph $G_q=(S_q,E_q,w_q)$, where $|S_q|=k$. They then refine their responses
through forward and reverse message passing over the graph. Finally, a
meta-agent uses mean pooling to synthesize the refined responses and relevance
scores into the final prediction.

\section{Experiments} 

\vspace{-0.5em}

In this section, we conduct extensive experiments to systematically investigate multi-agent coordination for graph learning through the following research questions (\textbf {RQs}): \textbf{RQ1:} How effective is multi-agent graph learning compared with single-agent reasoning? \textbf{RQ2:} What drives the gains of multi-agent graph learning? \textbf{RQ3:} How do different coordination strategies compare in effectiveness and efficiency? \textbf{RQ4:} How does learning affect multi-agent graph learning?

\vspace{-0.5em}

\paragraph{Roadmap.} We compare multi-agent and single-agent methods across backbones (\S\ref{sec:rq1}; App.~\ref{app:baseline-variants},~\ref{app:backbone_ablation}), identify sources of multi-agent gains across coordination methods (\S\ref{sec:rq2}; App.~\ref{app:rq2}--~\ref{app:decomposition}), evaluate coordination effectiveness and efficiency under specialist disagreement (\S\ref{sec:rq3}; App.~\ref{app:agreement},~\ref{app:computational-cost}), and study learned coordination (\S\ref{sec:rq4}). The appendix reports dataset statistics (App.~\ref{app:datasets}), implementation details (App.~\ref{app:implementation}), oracle gaps (App.~\ref{app:oracle}), and prompts (App.~\ref{app:prompts}).

\vspace{-0.5em}

\subsection{Experimental Setup}
\label{sec:exp-setup}
\vspace{-0.5em}

\paragraph{Datasets.}
We evaluate on Node Classification (NC) and Link Prediction (LP) across \textbf{7} text-attributed graph (TAG) benchmarks spanning \textbf{3} domains: \textbf{1) citation networks}: ogbn-Arxiv~\citeyearpar{hu2020open},
PubMed~\citeyearpar{sen2008collective}, and Arxiv-2023~\citeyearpar{he2024harnessing};
\textbf{2) e-commerce networks}: ogbn-Products~\citeyearpar{hu2020open},
Amazon-Sports~\citeyearpar{yan2023comprehensive}, and
Amazon-Computers~\citeyearpar{yan2023comprehensive}; and \textbf{3) social networks}: Reddit~\citeyearpar{yan2025graphmeetsmultimodal}.
Dataset statistics and task construction are provided in
Appendix~\ref{app:datasets}.

\vspace{-0.5em}

\paragraph{Baselines.}
We compare the multi-agent coordination methods instantiated in GraphMAS against representative baselines spanning direct LLM reasoning and LLM-based graph learning.
\textbf{1) LLM reasoning.}
\emph{Chain-of-Thought (CoT)}~\citeyearpar{wei2022cot} predicts directly without access
to graph information.
\textbf{2) Agentic methods.}
We include \emph{Search-o1}~\citeyearpar{li2025searcho1}, a general-purpose agentic
retrieval method, together with \emph{Graph-CoT}~\citeyearpar{jin2024graphcot} and
\emph{GraphSearch}~\citeyearpar{liu2026graphsearch}, which explicitly interact with graph
information.
Beyond these baselines, we evaluate all four GraphMAS specialists individually and systematically compare the seven multi-agent coordination
methods introduced in Section~\ref{sec:methods}:
\emph{Majority Vote}~\citeyearpar{li2024more},
\emph{LLM Aggregation}~\citeyearpar{jiang2023llmblender},
\emph{SOP Workflow}~\citeyearpar{hong2024metagpt},
\emph{Multi-Agent Debate}~\citeyearpar{du2024debate},
\emph{Adaptive Routing}~\citeyearpar{zhang2025routerr1},
\emph{DyLAN}~\citeyearpar{liu2024dylan}, and
\emph{Graph of Agents}~\citeyearpar{yun2026goa}.

\vspace{-0.5em}


\subsection{Performance of Multi-Agent Graph Learning (RQ1)} 
\label{sec:rq1}
\label{sec}
\vspace{-0.5em}

\begin{table*}[t]
\vspace{-2em}
\centering
\caption{
Accuracy (\%) of LLM reasoning, agentic baselines, graph specialists, and multi-agent methods, with all LLM-based methods using Qwen2.5-32B-Instruct. Avg.\ averages seven datasets; Avg.\ Rank averages ranks across 15 methods and 14 dataset--task settings. Top 2 counts settings reaching either of the two highest distinct accuracies. Best and second-best values are {\setlength{\fboxsep}{1pt}\colorbox{bestcolor}{\textbf{bold}}} and {\setlength{\fboxsep}{1pt}\colorbox{secondcolor}{\underline{underlined}}}, respectively. See App.~\ref{app:implementation} for implementation details.
}
\label{tab:main_results}

\setlength{\tabcolsep}{2.6pt}
\renewcommand{\arraystretch}{1.18}
\setlength{\aboverulesep}{0pt}   
\setlength{\belowrulesep}{0pt}

\resizebox{\textwidth}{!}{%
\begin{tabular}{
@{}l|
*{7}{c}>{\columncolor{summaryblue}}c|
*{7}{c}>{\columncolor{summaryblue}}c|
cc@{}
}
\toprule[1.2pt]

\rowcolor{headerbg}
\textbf{Method}
& \multicolumn{8}{c|}{\textbf{Link Prediction}}
& \multicolumn{8}{c|}{\textbf{Node Classification}}
& \multicolumn{2}{c}{\textbf{Overall}} \\

\cmidrule(lr){2-9}
\cmidrule(lr){10-17}
\cmidrule(lr){18-19}

\rowcolor{headerbg}
& \textbf{arXiv} & \textbf{products} & \textbf{PubMed} & \textbf{Reddit}
& \textbf{sports} & \textbf{arXiv-23} & \textbf{computers} & \textbf{Avg.}
& \textbf{arXiv} & \textbf{products} & \textbf{PubMed} & \textbf{Reddit}
& \textbf{sports} & \textbf{arXiv-23} & \textbf{computers} & \textbf{Avg.}
& \textbf{Avg. Rank} & \textbf{Top 2} \\

\midrule[0.8pt]

\rowcolor{groupgray}
\multicolumn{19}{c}{\textbf{LLM Reasoning}} \\

\textit{Chain-of-Thought}
& 76.1 & 63.8 & 80.4 & 55.3 & 64.8 & 78.0 & 63.7 & 68.9
& 56.5 & 66.1 & 91.6 & 60.1 & 53.8 & 60.1 & 59.4 & 63.9
& 10.8 & 0 \\

\groupsep

\rowcolor{groupgray}
\multicolumn{19}{c}{\textbf{Agentic Methods}} \\

\textit{Search-o1}
& 63.8 & 79.5 & 69.8 & 64.7 & 79.1 & 65.3 & 66.6 & 69.8
& 57.3 & 68.1 & 89.1 & 62.0 & 59.7 & 60.4 & 60.9 & 65.4
& 11.0 & 0 \\

\textit{Graph-CoT}
& 60.2 & 78.6 & 60.9 & 60.4 & 70.8 & 62.0 & 62.2 & 65.0
& 54.5 & 64.6 & 70.5 & 56.1 & 57.2 & 59.7 & 62.6 & 60.7
& 13.7 & 0 \\

\textit{GraphSearch}
& 70.3 & 79.5 & 75.2 & 62.7 & 74.5 & 70.5 & 68.3 & 71.6
& 57.6 & 71.7 & 89.8 & 67.4 & 59.8 & 55.8 & \best{69.9} & 67.4
& 9.4 & 1 \\

\groupsep

\rowcolor{groupgray}
\multicolumn{19}{c}{\textbf{Specialist Agents}} \\

\textit{Proximal Neighborhood}
& 75.4 & 76.7 & 74.8 & 67.1 & 78.8 & 71.6 & 71.6 & 73.7
& 56.1 & 70.4 & 90.7 & \best{68.3} & 60.1 & 58.3 & 69.3 & 67.6
& 8.4 & 1 \\

\textit{Distal Neighborhood}
& 72.2 & 83.2 & 71.0 & 50.7 & 75.5 & 72.1 & 68.1 & 70.4
& 57.7 & 71.8 & 90.5 & 58.1 & \best{62.8} & 58.0 & 63.6 & 66.1
& 10.3 & 1 \\

\textit{Global Relevance}
& 68.2 & 82.7 & 76.4 & 58.7 & 78.6 & 69.9 & 72.8 & 72.5
& 57.5 & 72.8 & 91.1 & 65.9 & 59.6 & 57.5 & 69.3 & 67.7
& 8.8 & 0 \\

\textit{Semantic Affinity}
& 72.6 & 82.6 & \best{81.2} & 60.8 & 80.7 & 73.5 & 68.8 & 74.3
& 59.4 & 70.6 & 90.5 & 61.2 & 58.5 & 59.3 & 62.3 & 66.0
& 9.2 & 1 \\

\groupsep

\rowcolor{groupgray}
\multicolumn{19}{c}{\textbf{General Multi-Agent Methods}} \\

\rowcolor{subgroupgray}
\multicolumn{19}{c}{\textit{Static Independent}} \\

\textit{Majority Vote}
& 74.6 & 84.3 & \second{81.0} & 61.0 & 84.8 & 71.5 & \second{74.5} & 76.0
& 58.7 & \second{73.2} & 91.6 & 67.1 & 61.4 & 59.9 & 68.8 & 68.7
& 5.8 & 3 \\

\textit{LLM Aggregation}
& \second{82.9} & 88.8 & 75.6 & \best{69.0} & \second{87.2} & \second{81.7} & \best{74.9} & \second{80.0}
& 59.3 & 71.9 & \best{92.2} & 67.0 & 59.2 & \second{61.0} & 69.3 & 68.6
& \second{3.9} & \best{7} \\

\subsep

\rowcolor{subgroupgray}
\multicolumn{19}{c}{\textit{Static Interactive}} \\

\textit{SOP Workflow}
& 73.5 & 86.5 & 71.3 & 62.3 & 81.8 & 75.7 & 71.0 & 74.6
& 56.6 & \best{73.8} & 90.1 & 66.1 & 60.4 & 57.4 & 69.2 & 67.7
& 8.1 & 1 \\

\textit{Multi-Agent Debate}
& 72.8 & 87.7 & 75.8 & 58.5 & 86.3 & 73.9 & 74.0 & 75.6
& 59.4 & 71.7 & 91.1 & 67.5 & 58.2 & 59.8 & 69.0 & 68.1
& 7.1 & 0 \\

\subsep

\rowcolor{subgroupgray}
\multicolumn{19}{c}{\textit{Dynamic Independent}} \\

\textit{Adaptive Routing}
& \best{83.5} & \best{89.8} & 76.7 & \second{67.2} & 86.9 & \best{83.3} & 74.3 & \best{80.2}
& \best{60.1} & 72.7 & 92.0 & \second{67.7} & 59.0 & \second{61.0} & 68.9 & \second{68.8}
& \best{3.3} & \best{7} \\

\subsep

\rowcolor{subgroupgray}
\multicolumn{19}{c}{\textit{Dynamic Interactive}} \\

\textit{DyLAN}
& 79.7 & 86.9 & 73.7 & 58.1 & 86.4 & 76.2 & 72.7 & 76.2
& \second{60.0} & 72.5 & 91.5 & 67.3 & \second{61.5} & \best{61.3} & \second{69.8} & \best{69.1}
& 5.1 & \second{4} \\

\textit{Graph of Agents}
& 82.3 & \second{89.4} & 75.7 & 63.0 & \best{87.4} & 79.8 & 74.0 & 78.8
& 59.5 & 71.4 & \second{92.1} & 66.5 & 58.9 & \second{61.0} & 68.3 & 68.2
& 5.2 & \second{4} \\

\bottomrule[1.2pt]
\end{tabular}%
}
\vspace{-0.7em}
\end{table*}

Table~\ref{tab:main_results} compares direct LLM reasoning, agentic graph reasoning, individual graph specialists, and multi-agent coordination across seven datasets and both graph learning tasks.

\noindent\textbf{\textit{$\filledstar$ Finding 1: Graph specialists are individually competitive and exhibit distinct strengths across datasets and tasks.}} The specialist agents already provide strong standalone graph reasoning.
The best specialist-level average reaches \textbf{74.3\%} on LP and \textbf{67.7\%} on NC, compared with 71.6\% and 67.4\% for GraphSearch, respectively. More importantly, the strongest specialist varies across datasets and tasks,
indicating that different graph evidence perspectives capture complementary
predictive signals rather than a universally superior graph view.

\noindent\textbf{\textit{$\filledstar$ Finding 2: Coordinating graph specialists consistently improves over individual specialist reasoning, especially on link prediction.}}
On LP, all seven multi-agent methods outperform the strongest individual
specialist on average, with the best result improving from \textbf{74.3\% to
80.2\%}. On NC, six methods improve, and one matches the strongest specialist,
raising the best average from \textbf{67.7\% to 69.1\%}. Multi-agent methods
also achieve better overall ranks, ranging from \textbf{3.3 to 8.1}, compared
with 8.4 for the strongest single-agent method. This shows that
coordinating specialist judgments provides gains beyond individual specialists
and conventional single-agent reasoning.

\vspace{-0.5em}

\subsection{Sources of Multi-Agent Gains (RQ2)}
\label{sec:rq2}
\vspace{-0.5em}

\begin{wraptable}{r}{0.35\columnwidth}
\vspace{-2.5em}
\centering
\caption{
Multi-agent gains under \emph{LLM Aggregation}. \cmark/\xmark\ denote
specialized/shared evidence or roles. Results are average accuracy (\%).
}
\label{tab:rq2}
\small
\setlength{\tabcolsep}{5pt}
\begin{tabular}{cccc}
\toprule
\textbf{Evidence}
& \textbf{Role}
& \textbf{LP} & \textbf{NC} \\
\midrule

\multicolumn{4}{l}{\textit{(a) Specialist Design}} \\
\xmark & \xmark & 73.0 & 66.0 \\
\xmark & \cmark & 69.9 & 65.4 \\
\cmark & \xmark & 73.7 & 66.1 \\
\cmark & \cmark & \textbf{80.0} & \textbf{67.7} \\

\midrule

\multicolumn{4}{l}{\textit{(b) Reasoning decomposition}} \\
\multicolumn{2}{l}{Single-agent reasoning}
& 74.6 & 65.6 \\
\multicolumn{2}{l}{Multi-agent reasoning}
& \textbf{80.0} & \textbf{67.7} \\

\bottomrule
\end{tabular}

\end{wraptable}

We examine two sources of multi-agent gains: \textbf{specialist design},
which varies evidence and reasoning roles, and \textbf{reasoning decomposition},
which compares joint versus separate reasoning over complementary graph
perspectives. To isolate these effects, we use the non-interactive
\emph{LLM Aggregation}. See Appendix~\ref{app:rq2}--\ref{app:decomposition} for details and additional results.

\vspace{-0.4em}

\noindent\textbf{\textit{$\filledstar$ Finding 3: Agent-specific evidence and specialized roles are most effective when combined.}}
We vary whether agents receive distinct graph evidence and specialized roles,
while fixing the number of agents, backbone, and coordination mechanism.
As shown in Table~\ref{tab:rq2}(a), specialized roles alone do not help with
shared evidence, reducing LP accuracy from 73.0\% to 69.9\%, while
agent-specific evidence with generic roles yields only marginal gains.
Combining both gives the strongest results, reaching \textbf{80.0\%} on LP
and \textbf{67.7\%} on NC. These results suggest that effective specialization arises from aligning distinct graph evidence with each agent's reasoning role, rather than from either component in isolation. 

\vspace{-0.4em}

\noindent\textbf{\textit{$\filledstar$ Finding 4: Reasoning decomposition provides gains beyond access to diverse graph evidence.}}
We next examine whether reasoning decomposition provides benefits by comparing a multi-agent system with a single agent that receives the same four evidence sources jointly in one reasoning context.
As shown in Table~\ref{tab:rq2}(b), single-agent reasoning achieves 74.6\% on LP and 65.6\% on NC, while multi-agent reasoning reaches \textbf{80.0\%} and \textbf{67.7\%}. Thus, the gains extend beyond broader evidence access to decomposing complementary perspectives across separate reasoning processes.

\vspace{-0.5em}
\subsection{Effectiveness and Efficiency of Coordination (RQ3)}
\label{sec:rq3}

\vspace{-0.5em}

\begin{wrapfigure}{r}{0.4\linewidth}
    \vspace{-1em}
    \centering
    \includegraphics[width=\linewidth]{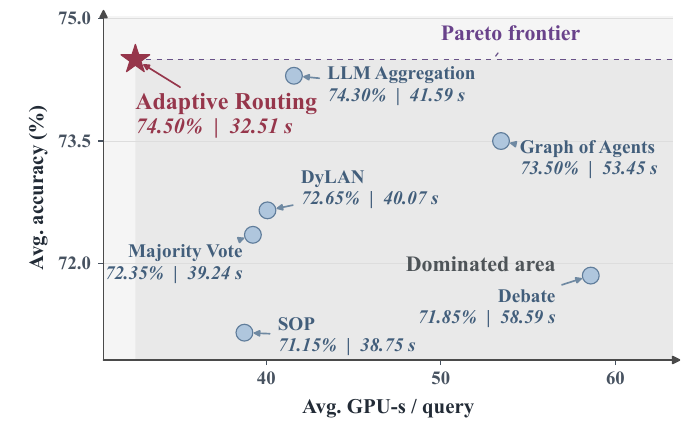}
    \vspace{-1.5em}
\caption{Accuracy--cost trade-offs averaged across NC and LP.}
    \label{fig:pareto}
\end{wrapfigure}

We compare the seven GraphMAS coordination methods along two questions:
which coordination designs provide the strongest predictive and computational
trade-offs, and under what conditions their performance differs most.

\vspace{-0.4em}

\noindent\textbf{\textit{$\filledstar$ Finding 5: Instance-adaptive specialist selection is more effective than simply increasing inter-agent interaction.}} Despite sharing the same specialist pool, different coordination paradigms
exhibit substantial performance differences. Among them, \emph{Adaptive Routing} achieves the best overall average rank of \textbf{3.3} and the highest LP accuracy of \textbf{80.2\%}, while \emph{DyLAN} obtains the highest NC accuracy of \textbf{69.1\%}.
Interestingly, the simpler static-independent \emph{LLM Aggregation} remains highly competitive, reaching \textbf{80.0\%} on LP with an average rank of 3.9.
In contrast, the static-interactive \emph{SOP Workflow} and \emph{Multi-Agent Debate} do not consistently outperform independent coordination.

Figure~\ref{fig:pareto} further compares the accuracy--efficiency trade-off across coordination paradigms.
Overall, \emph{Adaptive Routing} is the only method on the Pareto frontier, achieving strong predictive performance with lower inference cost than methods that invoke the full specialist pool.
By contrast, interaction-heavy methods such as \emph{SOP Workflow}, \emph{Multi-Agent Debate}, and \emph{Graph of Agents} incur higher computation without corresponding accuracy improvements. Detailed task-specific results and statistics are reported in Appendix~\ref{app:computational-cost}. 

\begin{wrapfigure}{r}{0.4\linewidth}
    \vspace{-1em}
    \centering
    \includegraphics[width=\linewidth]{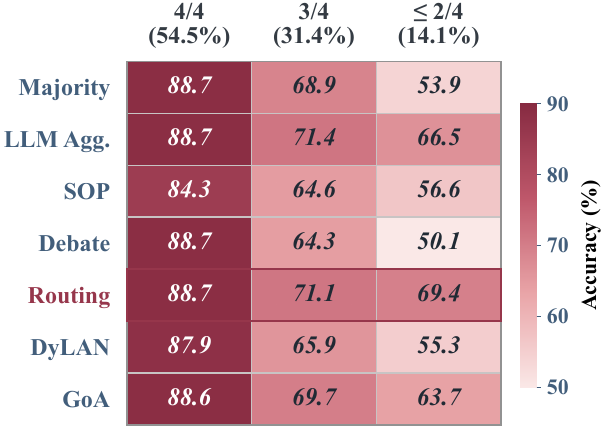}
    \vspace{-1.5em}
    \caption{LP accuracy (\%) by initial specialist agreement. Parentheses show each group's share.}
    \vspace{-2em}
    \label{fig:agreement_lp}
\end{wrapfigure}


\noindent\textbf{\textit{$\filledstar$ Finding 6: Coordination strategies differentiate most strongly on instances with specialist disagreement.}}
We stratify instances by agreement among the four initial specialist
predictions. On LP, the accuracy range across methods grows from 4.4 points under unanimous agreement to 7.1 points in the $3/4$ group and 19.3 points in the $\leq 2/4$ group, where \emph{Adaptive Routing} reaches 69.4\% while \emph{Multi-Agent Debate} obtains 50.1\%. NC follows the same trend
at a smaller scale, with the range increasing from 1.0 to 5.0 points
(Appendix~\ref{app:agreement}). Thus, disagreement does not by itself
guarantee coordination gains; rather, it is where the choice of coordination
mechanism matters most.

\vspace{-0.5em}
\subsection{Extending GraphMAS to Learned Coordination (RQ4)}
\label{sec:rq4}
\vspace{-0.5em}

We further study whether \textbf{learning the coordination policy} improves
multi-agent graph reasoning. Since \emph{Adaptive Routing} offers the strongest
accuracy--cost trade-off among training-free methods, we optimize its router
with PPO~\citep{schulman2017proximalpolicyoptimizationalgorithms} while keeping
the four graph specialists frozen. We compare the learned router with its
training-free counterpart and trained baselines, including GNN, LLM-based, and agentic graph-learning
baselines, under a shared train--transfer split. Separate NC and LP routers are
trained on four datasets and evaluated on seen and three held-out graphs.
Training details and baseline protocols are provided in
Appendices~\ref{app:router_training} and~\ref{app:learned_baselines}.

\begin{table*}[h]
\centering
\vspace{-0.8em}
\caption{
Accuracy (\%) of learned graph-learning methods under a shared
in-domain/transfer split. All LLM-based baselines use Qwen2.5-7B-Instruct; GraphMAS uses a 7B router with four frozen Qwen2.5-32B-Instruct specialists. In-domain/Transfer denote seen/unseen datasets during training. Our method is {\setlength{\fboxsep}{1pt}\colorbox{secondcolor}{highlighted}}.
Best and second-best results are \textbf{bold} and \underline{underlined}.
}
\label{tab:learned_magl}

\setlength{\tabcolsep}{2.35pt}
\renewcommand{\arraystretch}{1.18}
\setlength{\aboverulesep}{0pt}
\setlength{\belowrulesep}{0pt}

\resizebox{\textwidth}{!}{%
\begin{tabular}{
@{}ll|
*{7}{c}>{\columncolor{summaryblue}}c|
*{7}{c}>{\columncolor{summaryblue}}c
@{}
}
\toprule[1.2pt]

\rowcolor{headerbg}
&
& \multicolumn{8}{c|}{\textbf{Node Classification}}
& \multicolumn{8}{c}{\textbf{Link Prediction}} \\

\cmidrule(lr){3-10}
\cmidrule(lr){11-18}

\rowcolor{headerbg}
&
& \multicolumn{4}{c}{\textit{In-domain}}
& \multicolumn{3}{c}{\textit{Transfer}}
&
& \multicolumn{4}{c}{\textit{In-domain}}
& \multicolumn{3}{c}{\textit{Transfer}}
& \\

\cmidrule(lr){3-6}
\cmidrule(lr){7-9}
\cmidrule(lr){11-14}
\cmidrule(lr){15-17}

\rowcolor{headerbg}
\multirow{-3}{*}{\textbf{Category}}
& \multirow{-3}{*}{\textbf{Method}}
& arXiv & products & PubMed & computers
& Reddit & sports & arXiv-23
& \multirow{-2}{*}{\textbf{Avg.}}
& arXiv & products & PubMed & computers
& Reddit & sports & arXiv-23
& \multirow{-2}{*}{\textbf{Avg.}} \\

\midrule[0.8pt]

\multirow{3}{*}{GNN}
& \textit{GCN}
& 53.2 & 62.5 & 73.9 & 69.4 & 4.1 & 4.4 & 1.8 & 38.5
& 74.6 & 59.9 & 49.8 & 72.4 & 50.1 & 49.9 & 54.8 & 58.8 \\

& \textit{RevGAT}
& 47.1 & 48.4 & 81.5 & 68.8 & 3.9 & 17.5 & 0.7 & 38.3
& 73.6 & 57.1 & 58.6 & 69.9 & 49.9 & 49.3 & 49.9 & 58.3 \\

& \textit{GraphSAGE}
& 62.0 & 60.4 & 51.6 & \textbf{75.2} & 5.9 & 6.6 & 1.6 & 37.6
& 71.1 & 77.2 & 78.1 & 70.3 & 55.0 & 52.1 & 47.5 & 64.5 \\

\subsep

\multirow{2}{*}{LLM-based GL}
& \textit{GraphPrompter}
& 55.2 & 68.8 & 90.3 & 60.6 & 54.8 & 30.6 & 23.6 & 54.8
& \underline{91.7} & \underline{91.4} & \underline{87.5} & \textbf{79.9}
& 61.5 & 85.2 & \underline{89.2} & \underline{83.8} \\

& \textit{LLaGA}
& 64.2 & 56.4 & 86.5 & 72.1 & 7.8 & 0.3 & 18.8 & 43.7
& 79.5 & 68.1 & 81.9 & 65.6 & 59.9 & 60.2 & 62.5 & 68.2 \\

\subsep

Agentic GL
& \textit{AgentGL-7B-GRPO}
& \underline{67.8} & \textbf{75.3} & \underline{92.2} & 64.8
& 43.2 & 55.6 & \underline{69.8} & 67.0
& 87.3 & 91.1 & 86.7 & 74.8
& \underline{72.8} & 82.6 & 87.4 & 83.2 \\

\groupsep

\rowcolor{secondcolor}
& \textit{Routing} (Training-free)
& 59.1 & 72.9 & 90.7 & 68.3
& \underline{67.5} & \underline{59.9} & 60.5 & \underline{68.4}
& 83.5 & 89.4 & 76.2 & 72.8
& 64.7 & \underline{87.4} & 83.3 & 79.6 \\

\rowcolor{secondcolor}
\multirow{-2}{*}{\textbf{GraphMAS}}
& \textit{Routing} (PPO)
& \textbf{68.4} & \underline{73.9} & \textbf{93.1} & \underline{72.2}
& \textbf{68.0} & \textbf{61.3} & \textbf{70.0} & \textbf{72.4}
& \textbf{94.8} & \textbf{93.6} & \textbf{90.8} & \underline{76.2}
& \textbf{73.3} & \textbf{89.1} & \textbf{93.0} & \textbf{87.3} \\

\bottomrule[1.2pt]
\end{tabular}%
}

\vspace{-0.3em}
\end{table*}

\noindent\textbf{\textit{$\filledstar$ Finding 7: Learning the coordination
policy improves multi-agent graph learning with fixed specialists.}}
PPO raises average LP accuracy from \textbf{79.6\% to 87.3\%}. On NC, the
learned router achieves \textbf{72.4\%}, compared with 68.4\% for its
training-free counterpart. Because the specialists remain frozen, these gains reflect improved
coordination rather than changes in specialist capability.

\noindent\textbf{\textit{$\filledstar$ Finding 8: Learned coordination
generalizes beyond router-training graphs.}}
The learned router improves on both training and held-out datasets. On LP, it
raises average accuracy on Reddit, Sports, and arXiv-23 from
\textbf{78.5\% to 85.1\%} and outperforms AgentGL-7B-GRPO on all seven
datasets. These results show that the learned specialist-selection policy
extends beyond the training graphs and provides a strong learned-coordination
baseline for GraphMAS.

\vspace{-0.5em}

\section{Conclusion}

\vspace{-0.5em}

We presented \textsc{GraphMAS}, a controlled framework for studying multi-agent coordination in graph learning. Our results suggest that the value of multi-agent reasoning lies less in simply adding more agents or communication, and more in how complementary graph perspectives are decomposed, specialized, and selectively coordinated. Across tasks and coordination paradigms, carefully structured specialist reasoning can outperform both individual specialists and strong single-agent systems, while adaptive coordination can achieve competitive performance with substantially less specialist computation. Moreover, further gains from learning suggest that coordination itself can be a learnable component of graph reasoning systems. These findings position multi-agent graph learning as a problem of coordination design rather than agent scaling alone. We hope \textsc{GraphMAS} provides a common foundation for developing coordination mechanisms that more effectively exploit complementary graph evidence across diverse graph reasoning settings.

\subsection*{AI use statement}

In this work, we used generative AI tools solely for proofreading and improving the readability of the manuscript. We did not use generative AI to clean or reformat datasets, conduct qualitative or thematic analyses, or interpret results. The remaining disclosure categories do not apply to this work. We independently reviewed the final manuscript to ensure that it accurately reflects our intended meaning. We take full responsibility for all content in this work, including any text, claims, or artifacts produced with the assistance of generative AI.

\subsection*{Reproducibility statement}

We have made efforts to ensure the reproducibility of our results. The benchmark code is available at the anonymous repository linked in the abstract. Section~\ref{sec:specialists} describes the graph reasoning specialists and their shared interface, and Section~\ref{sec:methods} and Table~\ref{tab:paradigms} specify all seven coordination methods. Appendix~\ref{app:datasets} describes the datasets, preprocessing, evaluation sampling, and link prediction construction. Appendix~\ref{app:general} reports the backbone models, decoding settings, retrieval configurations, and coordination hyperparameters; Appendix~\ref{app:router_training} details learned-router training, with hyperparameters in Table~\ref{tab:router_hparams}; and Appendix~\ref{app:learned_baselines} describes the baseline protocols. Appendix~\ref{app:prompts} provides the core prompt templates for all specialists and coordination methods. All models are publicly available Hugging Face checkpoints, explicitly mentioned in Appendix~\ref{app:implementation}.

\bibliography{iclr2027_conference}
\bibliographystyle{iclr2027_conference}

\appendix
\section{Dataset Details}
\label{app:datasets}

We evaluate on seven text-attributed graph (TAG) benchmarks spanning three
domains: citation networks, e-commerce graphs, and social networks.
Each node is associated with
natural-language attributes, such as paper titles and abstracts, product
descriptions, or social-media text, while edges encode the native relations
of the corresponding graph. These benchmarks therefore require reasoning
jointly over node semantics and graph structure. We consider both node
classification (NC), which predicts the original multi-class node label, and
link prediction (LP), which predicts whether an edge exists between a pair
of nodes. We follow the dataset splits aligned with past research ~\citep{sun2025graphicl,sun2026agentgl,liu2026graphsearch} and use the official test split when available. For Reddit, which originates from a multimodal graph benchmark, we remove image attributes and retain only textual node attributes, yielding a TAG setting. We provide the same processed graph and textual attributes to every method. Table~\ref{tab:dataset_statistics} summarizes the statistics of TAG datasets used in our experiments.

Following prior agentic graph learning research \citep{liu2026graphsearch,sun2026agentgl}, we subsample the evaluation sets to control inference
cost. For each dataset--task pair, we randomly sample 1,000 instances
from its designated test set, where an instance is a target node for NC and a
node pair for LP. For LP, evaluation sets are balanced within each dataset, with 500 positive pairs, sampled from held-out edges, and 500 negative pairs, sampled uniformly from unconnected node pairs.
All evaluation edges are removed from the graph exposed to specialists and baselines, so retrieval cannot directly reveal the queried link. When fewer than 1,000 eligible test instances are available,
we evaluate on the full available set. The resulting sampled instances are
fixed across all compared methods to ensure that performance and inference
cost are measured on identical queries.

\begin{table*}[h]
\centering
\caption{
Statistics of the seven text-attributed graphs used in GraphMAS.
\#Classes denotes the node-classification label space; link prediction is
formulated as binary prediction.
}
\small
\setlength{\tabcolsep}{7pt}
\renewcommand{\arraystretch}{1.10}
\begin{tabular}{llrrr}
\toprule
\textbf{Domain}
& \textbf{Dataset}
& \textbf{\#Nodes}
& \textbf{\#Edges}
& \textbf{\#Classes} \\
\midrule

\multirow{4}{*}{Citation Network}
& ogbn-Arxiv       & 169,343 & 1,166,245 & 40 \\
& PubMed          & 19,717  & 44,338    & 3  \\
& Arxiv-2023      & 46,198  & 78,548    & 40 \\
\midrule

\multirow{3}{*}{E-commerce}
& ogbn-Products (subset) & 54,025  & 74,420    & 47 \\
& Amazon-Sports         & 173,055 & 1,946,555 & 13 \\
& Amazon-Computers      & 87,229  & 808,310   & 10 \\
\midrule

Social Network
& Reddit         & 13,037  & 566,160   & 20 \\
\bottomrule
\end{tabular}

\label{tab:dataset_statistics}
\end{table*}

\section{Implementation Details and Hyperparameters}
\label{app:implementation}
\subsection{General and Training-Free Methods}
\label{app:general}
 Following prior work~\citep{chen2024llaga,sun2026agentgl,sun2025graphicl, jin2024graphcot,liu2026graphsearch}, we use \textbf{accuracy} as the primary metric for all graph reasoning tasks. All results are reported from single runs conducted under the same experimental setup on 2 * NVIDIA A100 80 GB GPUs. We follow standard training and evaluation protocols for all baselines~\citep{chen2024llaga,10.1145/3589335.3651476}.

For a controlled comparison across LLM-based graph learning methods and multi-agent systems, we use the same official Hugging Face checkpoint, \texttt{Qwen/Qwen2.5-32B-Instruct}~\citep{qwen2024qwen25}, for training-free methods. We use a decoding temperature of $0.7$ for all Qwen2.5 inference
calls, including those made by baselines, graph
specialists, and coordination modules.

For Proximal and Distal Neighborhood retrieval, we cap the candidate set at $200$ nodes, augment each search query with the target node's title for candidate ranking, and return the top $3$ nodes as evidence.
Semantic affinity retrieval uses the SentenceTransformer model
\texttt{all-mpnet-base-v2}~\citep{reimers2019sentencebert,song2020mpnet}.
We precompute 768-dimensional corpus embeddings and rank candidates by
cosine similarity between $\ell_2$-normalized representations. For each
query node, the semantic neighbor list contains the top $10$ nodes after
excluding the query node itself, of which the first $3$ are provided to
the agent.
For the Global Relevance specialist, we approximate personalized
PageRank (PPR) using APPNP~\citep{gasteiger2018combining} with
$\alpha=0.15$ and $K=30$. Here, $\alpha$ denotes the teleport probability
to the personalization seed at each propagation step, and $K$ is the
number of propagation iterations. The resulting scores determine the
PPR-ranked node lists.

Each specialist may issue at most 4 \texttt{<search>} calls per instance, after which it must commit to a prediction.
Adaptive Routing may invoke at most $M{=}4$ specialist calls before final synthesis, matching the learned router.
Multi-Agent Debate runs at most $R{=}2$ revision rounds.
DyLAN and Graph of Agents both retain the top $k{=}3$ specialists after ranking and relevance scoring, respectively.

\subsection{Training Details of the Learned Router}
\label{app:router_training}

\paragraph{Routing formulation.}
We train only the router agent of \emph{Adaptive Routing}, while keeping the
four graph specialists frozen. Each episode corresponds to one prediction
instance (a node for NC or a node pair for LP). At each turn, the router
observes the target information, label space, previously collected specialist
reports, and remaining call budget, and chooses either to invoke one specialist
or to produce the final prediction. The router can make at most four specialist
calls and must invoke at least one specialist before producing the final answer.

\paragraph{Reward.}
We optimize a terminal reward that jointly encourages prediction accuracy and
efficient specialist usage:
\begin{equation}
r =
\begin{cases}
r_{\mathrm{fmt}}, & \text{if the output format is invalid},\\[2pt]
(1-\alpha)R_{\mathrm{out}} + \alpha R_{\mathrm{cost}},
& \text{otherwise},
\end{cases}
\end{equation}
where $R_{\mathrm{out}}\in\{0,1\}$ denotes answer correctness and
\[
R_{\mathrm{cost}}
=
R_{\mathrm{out}}\cdot \frac{4-n_{\mathrm{calls}}}{3}.
\]

We reward call efficiency only on correct trajectories, preventing the router
from reducing cost through premature stopping at the expense of accuracy.

\paragraph{Training data.}
For each task, we sample 3{,}000 training instances per graph from the four in-domain graphs (arXiv, products, PubMed, computers), and hold
out a further 256 instances per graph for validation; both are disjoint from
the test instances. Dataset construction is detailed in
Appendix~\ref{app:datasets}. To make RL training tractable, each specialist's response to each training instance is precomputed once and replayed; at evaluation time, the frozen specialists are invoked online.

\paragraph{Optimization.}
The router is initialized from \texttt{Qwen/Qwen2.5-7B-Instruct}, which also serves as the KL-reference model, and is optimized with
PPO~\citep{schulman2017proximalpolicyoptimizationalgorithms} using a learned value head and GAE~\citep{schulman2018highdimensionalcontinuouscontrolusing}.
All specialists use Qwen2.5-32B-Instruct. NC and LP share the configuration in Table~\ref{tab:router_hparams}. Checkpoints are selected by accuracy on the in-domain validation split; the held-out transfer graphs are never used for training or model selection.

\begin{table}[h]
\centering
\small
\caption{Main hyperparameters for learned-router training.}
\label{tab:router_hparams}
\begin{tabular}{@{}ll@{}}
\toprule
\textbf{Hyperparameter} & \textbf{Value} \\
\midrule
Router initialization / reference & Qwen2.5-7B-Instruct \\
Frozen specialists & Qwen2.5-32B-Instruct \\
Training iterations & 150 \\
Trajectories per iteration & 128 \\
Maximum specialist calls & 4 \\
Actor / critic learning rate & $5\times10^{-7}$ / $1\times10^{-5}$ \\
PPO epochs & 2 \\
Mini-batch size & 32 \\
Policy / value clip ratio & 0.2 / 0.2 \\
KL coefficient & 0.002 \\
GAE $\gamma$ / $\lambda$ & 1.0 / 0.95 \\
Format reward $r_{\mathrm{fmt}}$ & $-1$ \\
Cost weight $\alpha$ & $0.1$ \\
\bottomrule
\end{tabular}
\end{table}

\subsection{Implementation of Learned Baselines}
\label{app:learned_baselines}

\paragraph{Shared protocol.}
Table~\ref{tab:learned_magl} compares learned GraphMAS with conventional GNN,
LLM-based graph-learning, and agentic baselines. All learned methods are
trained on the four in-domain graphs (arXiv, products, PubMed, computers).
\emph{In-domain} results are measured on test instances of these graphs that
are disjoint from training and validation; \emph{transfer} results are
measured on Reddit, sports, and arXiv-23, which are never seen during training
or model selection and receive no fine-tuning or adaptation. All methods are
evaluated on the same 1{,}000 test instances per dataset.

\paragraph{Backbone and compute.}
All LLM-based learned methods use Qwen2.5-7B-Instruct as the trainable
backbone. GraphMAS additionally queries four frozen Qwen2.5-32B-Instruct
specialists at inference time, so its trainable parameters match those of
the baselines but its inference-time model capacity does not. We therefore
treat the comparison against external baselines as a system-level reference,
and isolate the effect of router training with the controlled comparison
below.

\paragraph{Baseline Details.}
We include GCN~\citep{kipf2017semisupervised}, GraphSAGE~\citep{3294771.3294869}, and RevGAT~\citep{pmlr-v139-li21o} as conventional graph-learning
baselines. GraphPrompter~\citep{10.1145/3589335.3651476} and LLaGA~\citep{chen2024llaga} are trained and evaluated following their original
procedures under the shared split, on the same processed graph structures and
node texts used by GraphMAS. Since both predict labels in text, target label
names are provided in the prompt and no label-space mapping is needed. We further compare with AgentGL-7B-GRPO~\citep{sun2026agentgl} as a learned agentic graph-reasoning baseline. We retain its original training and inference procedure while using the same in-domain training graphs, held-out transfer graphs, and evaluation instances.

\paragraph{Controlled comparison: learned vs.\ training-free routing.}
Learned GraphMAS is compared with its training-free counterpart under
identical conditions: the same Qwen2.5-7B-Instruct router backbone, the same
four frozen specialists, the same retrieval settings, the same decoding
temperature, and the same test instances.

\section{Detailed Experiment Results}
\label{app:extra_exp}

\subsection{Specialist Complementarity and Oracle Gap}
\label{app:oracle}

Table~\ref{tab:specialist_oracle} details the performance of the best individual
specialist and the specialist oracle across all datasets. The average gap of \textbf{16.9 percentage points} on link prediction and \textbf{10.0 percentage points} on node classification
reveals substantial complementary strengths and motivates effective
coordination among specialists.

\begin{table}[t]
\centering
\caption{
Accuracy (\%) of the best individual specialist and the specialist oracle.
Best Specialist selects the highest-performing specialist separately for each
dataset, and Oracle counts an instance as correct if any specialist is correct.
$\Delta$ denotes the Oracle--Best Specialist gap in percentage points.
Avg.\ denotes the mean across the seven datasets.}
\label{tab:specialist_oracle}

\begin{tabular}{lccc|ccc}
\toprule
& \multicolumn{3}{c|}{\textbf{Link Prediction}}
& \multicolumn{3}{c}{\textbf{Node Classification}} \\
\cmidrule(lr){2-4}
\cmidrule(lr){5-7}
Dataset
& Best Specialist & Oracle & $\Delta$
& Best Specialist & Oracle & $\Delta$ \\
\midrule
arXiv
& 75.4 & \textbf{91.6} & $+16.2$
& 59.4 & \textbf{74.4} & $+15.0$ \\
products
& 83.2 & \textbf{96.0} & $+12.8$
& 72.8 & \textbf{80.6} & $+7.8$ \\
PubMed
& 81.2 & \textbf{98.2} & $+17.0$
& 91.1 & \textbf{94.8} & $+3.7$ \\
Reddit
& 67.1 & \textbf{83.5} & $+16.4$
& 68.3 & \textbf{74.8} & $+6.5$ \\
sports
& 80.7 & \textbf{97.0} & $+16.3$
& 62.8 & \textbf{75.3} & $+12.5$ \\
arXiv-23
& 73.5 & \textbf{92.4} & $+18.9$
& 59.3 & \textbf{72.1} & $+12.8$ \\
computers
& 72.8 & \textbf{93.3} & $+20.5$
& 69.3 & \textbf{81.0} & $+11.7$ \\
\midrule
Avg.
& 76.3 & \textbf{93.1} & $+16.9$
& 69.0 & \textbf{79.0} & $+10.0$ \\
\bottomrule
\end{tabular}

\end{table}

\subsection{Additional Results for Baseline Variants}
\label{app:baseline-variants}

GraphSearch~\citep{liu2026graphsearch} and Graph of
Agents~\citep{yun2026goa} each provide two variants.
GraphSearch-R expands neighborhoods recursively, whereas
GraphSearch-F flexibly retrieves across local and global neighborhoods.
GoA-mean aggregates refined responses, whereas
GoA-max selects the refined response of the most influential agent.
Table~\ref{tab:main_results} uses GraphSearch-F and GoA-mean. We ran these variants on the same datasets, and the results are detailed in Table~\ref{tab:baseline-variants}.

GraphSearch-R achieves lower average accuracy than GraphSearch-F
on both tasks and remains \textbf{below all evaluated multi-agent methods}
in average accuracy.
GoA-max improves average LP accuracy over GoA-mean from 78.8\%
to 79.7\%, approaching LLM Aggregation (80.0\%) and Adaptive
Routing (80.2\%), while both GoA variants achieve 68.2\% on NC.
However, \textbf{neither
alternative} exceeds the strongest multi-agent method in average
accuracy on either task.

\begin{table}[t]
\centering
\caption{Accuracy (\%) of baseline variants.
Avg.\ denotes the mean across the seven datasets.}
\label{tab:baseline-variants}
\resizebox{\textwidth}{!}{
\begin{tabular}{lcccccccc}
\toprule
Method & arXiv & products & PubMed & Reddit & sports
& arXiv-23 & computers & Avg. \\
\midrule
\multicolumn{9}{c}{\textbf{Link Prediction}} \\
\midrule
GraphSearch-R
& 63.8 & 82.2 & 70.6 & 67.8 & 76.7 & 63.5 & 67.2 & 70.3 \\
GoA-max
& 83.7 & 90.3 & 74.2 & 67.9 & 85.6 & 82.3 & 73.7 & 79.7 \\
\midrule
\multicolumn{9}{c}{\textbf{Node Classification}} \\
\midrule
GraphSearch-R
& 57.6 & 70.8 & 90.4 & 63.5 & 61.9 & 52.8 & 68.1 & 66.4 \\
GoA-max
& 59.3 & 71.4 & 92.2 & 66.2 & 59.2 & 60.9 & 68.3 & 68.2 \\
\bottomrule
\end{tabular}
}
\end{table}

\subsection{Detailed Results for Specialist Design}
\label{app:rq2}

We study specialist design through a controlled $2\times2$ ablation over
\textbf{evidence allocation} (shared vs.\ agent-specific) and
\textbf{role conditioning} (generic vs.\ specialized).
Under shared evidence, all four agents receive the same graph evidence,
collected from the four available perspectives within a fixed overall evidence
budget.
Under agent-specific evidence, the four agents receive different graph
views, with each agent restricted to one of the proximal neighborhood, distal neighborhood, graph-wide
relevance, and semantic-affinity perspectives.
Under generic roles, all agents follow the same general graph-reasoning
instruction, whereas under specialized roles, each agent is explicitly
conditioned to reason according to its assigned graph perspective.
The number of agents, backbone, evidence budget, and coordination mechanism
are held fixed across settings.
Table~\ref{tab:rq3_detailed} reports the specialist-design analysis across all
seven coordination methods, using 100 randomly sampled evaluation instances
per dataset and task.

\begin{table*}[h]
\centering
\caption{
Per-dataset accuracy (\%) under different specialist designs.
Shared/Agent-specific denote common/specialist-specific evidence;
Generic/Specialized denote general/specialist-specific role instructions.
\textbf{Bold} and \underline{underlined} values indicate the highest and
second-highest accuracies within each method and column.
}
\label{tab:rq3_detailed}
\setlength{\tabcolsep}{2.5pt}
\renewcommand{\arraystretch}{1.08}
\resizebox{\textwidth}{!}{%
\begin{tabular}{@{}lll|*{8}{c}|*{8}{c}@{}}
\toprule
\multirow{2}{*}{\textbf{Method}}
& \multirow{2}{*}{\textbf{Evidence}}
& \multirow{2}{*}{\textbf{Role}}
& \multicolumn{8}{c|}{\textbf{Link Prediction}}
& \multicolumn{8}{c}{\textbf{Node Classification}} \\
\cmidrule(lr){4-11}
\cmidrule(lr){12-19}
& &
& arXiv & products & PubMed & Reddit & sports & arXiv-23 & computers & Avg.
& arXiv & products & PubMed & Reddit & sports & arXiv-23 & computers & Avg. \\
\midrule

\multirow{4}{*}{\emph{Majority Vote}}
& Shared & Generic
& \underline{82} & 75 & 74 & \underline{58} & 78 & 74 & 71 & 73.1
& 63 & 68 & \textbf{92} & \underline{70} & 55 & 51 & 63 & 66.0 \\
& Shared & Specialized
& \textbf{83} & 77 & 70 & \textbf{62} & 76 & \underline{80} & 66 & 73.4
& 64 & 66 & \underline{91} & \textbf{72} & 55 & 51 & 62 & 65.9 \\
& Agent-specific & Generic
& \underline{82} & \underline{80} & \underline{78} & 56 & \underline{83} & \textbf{82} & \underline{76} & \underline{76.7}
& \underline{68} & \underline{69} & 89 & \underline{70} & \textbf{59} & \underline{54} & \underline{64} & \underline{67.6} \\
& Agent-specific & Specialized
& 80 & \textbf{85} & \textbf{80} & \textbf{62} & \textbf{85} & \textbf{82} & \textbf{80} & \textbf{79.1}
& \textbf{69} & \textbf{70} & 87 & \underline{70} & \underline{57} & \textbf{57} & \textbf{65} & \textbf{67.9} \\
\midrule

\multirow{4}{*}{\emph{LLM Aggregation}}
& Shared & Generic
& \underline{77} & 76 & \underline{72} & \underline{67} & 69 & \underline{82} & 68 & 73.0
& 61 & \underline{68} & 88 & \underline{70} & \underline{57} & \textbf{56} & \underline{62} & 66.0 \\
& Shared & Specialized
& 69 & 76 & 60 & 66 & \underline{73} & 72 & \underline{73} & 69.9
& 62 & 67 & \textbf{91} & 67 & \textbf{58} & 53 & 60 & 65.4 \\
& Agent-specific & Generic
& \textbf{87} & \underline{77} & 70 & 62 & 67 & \textbf{86} & 67 & \underline{73.7}
& \underline{65} & \underline{68} & \underline{89} & 68 & 55 & \textbf{56} & \underline{62} & \underline{66.1} \\
& Agent-specific & Specialized
& \textbf{87} & \textbf{85} & \textbf{77} & \textbf{68} & \textbf{89} & 77 & \textbf{77} & \textbf{80.0}
& \textbf{66} & \textbf{69} & \textbf{91} & \textbf{71} & \textbf{58} & \underline{54} & \textbf{65} & \textbf{67.7} \\
\midrule

\multirow{4}{*}{\emph{SOP Workflow}}
& Shared & Generic
& \underline{75} & 73 & \underline{66} & 50 & \underline{70} & \underline{78} & \underline{67} & \underline{68.4}
& \underline{57} & 66 & \textbf{89} & \underline{53} & 49 & \textbf{49} & \underline{55} & \underline{59.7} \\
& Shared & Specialized
& 58 & 63 & 48 & 47 & 59 & 64 & 55 & 56.3
& 51 & 64 & \underline{88} & 52 & 48 & 46 & 53 & 57.4 \\
& Agent-specific & Generic
& \textbf{82} & \underline{75} & 63 & \underline{55} & 61 & 73 & \underline{67} & 68.0
& 56 & \underline{67} & \textbf{89} & 35 & \underline{52} & \textbf{49} & 54 & 57.4 \\
& Agent-specific & Specialized
& 72 & \textbf{86} & \textbf{73} & \textbf{59} & \textbf{74} & \textbf{83} & \textbf{68} & \textbf{73.6}
& \textbf{62} & \textbf{68} & \underline{88} & \textbf{67} & \textbf{58} & \underline{47} & \textbf{57} & \textbf{63.9} \\
\midrule

\multirow{4}{*}{\emph{Multi-Agent Debate}}
& Shared & Generic
& \underline{88} & 77 & \underline{79} & 54 & \textbf{85} & 79 & 68 & 75.7
& \underline{65} & 61 & 91 & \textbf{73} & 51 & 53 & 63 & \underline{65.3} \\
& Shared & Specialized
& 83 & 80 & 76 & 54 & \underline{78} & 83 & 68 & 74.6
& 64 & 62 & \underline{92} & \underline{68} & 51 & 52 & \underline{68} & \underline{65.3} \\
& Agent-specific & Generic
& 83 & \textbf{82} & \textbf{82} & \underline{56} & 76 & \underline{86} & \underline{72} & \underline{76.7}
& \textbf{70} & \underline{64} & 91 & 60 & \underline{53} & \underline{54} & 63 & 65.0 \\
& Agent-specific & Specialized
& \textbf{89} & \underline{81} & 78 & \textbf{57} & \textbf{85} & \textbf{88} & \textbf{78} & \textbf{79.4}
& \textbf{70} & \textbf{67} & \textbf{93} & \textbf{73} & \textbf{54} & \textbf{55} & \textbf{70} & \textbf{68.9} \\
\midrule

\multirow{4}{*}{\emph{Adaptive Routing}}
& Shared & Generic
& \underline{87} & 77 & \underline{77} & 62 & \underline{85} & 81 & 70 & 77.0
& 66 & 65 & \textbf{91} & \textbf{71} & 55 & \textbf{55} & 60 & 66.1 \\
& Shared & Specialized
& \textbf{89} & 80 & 76 & \textbf{70} & 80 & \textbf{89} & 66 & 78.6
& 66 & 64 & \textbf{91} & \textbf{71} & 52 & 52 & \textbf{66} & 66.0 \\
& Agent-specific & Generic
& 83 & \underline{82} & \textbf{83} & 60 & 83 & \underline{86} & \underline{77} & \underline{79.1}
& \underline{71} & \textbf{69} & \underline{90} & \underline{69} & \underline{58} & \underline{54} & \underline{63} & \underline{67.7} \\
& Agent-specific & Specialized
& 86 & \textbf{84} & 76 & \underline{66} & \textbf{87} & 80 & \textbf{78} & \textbf{79.6}
& \textbf{73} & \underline{68} & \underline{90} & \textbf{71} & \textbf{60} & 52 & 61 & \textbf{67.9} \\
\midrule

\multirow{4}{*}{\emph{DyLAN}}
& Shared & Generic
& \textbf{82} & 75 & \underline{73} & \underline{57} & 78 & \underline{77} & 68 & 72.9
& 64 & 65 & \textbf{93} & \textbf{72} & 57 & \textbf{53} & 63 & 66.7 \\
& Shared & Specialized
& \textbf{82} & 77 & 69 & \textbf{61} & 75 & \underline{77} & 67 & 72.6
& \underline{66} & \underline{67} & 90 & \underline{71} & 54 & \textbf{53} & \underline{66} & 66.7 \\
& Agent-specific & Generic
& \underline{81} & \underline{78} & \textbf{83} & 54 & \underline{80} & \textbf{82} & \textbf{77} & \textbf{76.4}
& \textbf{69} & \textbf{70} & 89 & \textbf{72} & \underline{60} & \textbf{53} & 64 & \underline{68.1} \\
& Agent-specific & Specialized
& 79 & \textbf{82} & \textbf{83} & 52 & \textbf{86} & \underline{77} & \underline{75} & \underline{76.3}
& \textbf{69} & \underline{67} & \underline{91} & 70 & \textbf{61} & \textbf{53} & \textbf{67} & \textbf{68.3} \\
\midrule

\multirow{4}{*}{\emph{Graph of Agents}}
& Shared & Generic
& 82 & 73 & 76 & 58 & \underline{81} & 77 & 71 & 74.0
& 67 & 62 & \textbf{92} & \underline{71} & \underline{56} & \underline{53} & \underline{63} & 66.3 \\
& Shared & Specialized
& \underline{87} & 77 & 74 & \underline{61} & 77 & \textbf{85} & 69 & 75.7
& 65 & 62 & \underline{91} & \textbf{72} & 54 & \textbf{55} & 62 & 65.9 \\
& Agent-specific & Generic
& 82 & \underline{79} & \textbf{85} & 53 & 79 & \underline{81} & \underline{75} & \underline{76.3}
& \underline{70} & \underline{64} & 90 & 68 & \underline{56} & \textbf{55} & \textbf{64} & \underline{66.7} \\
& Agent-specific & Specialized
& \textbf{89} & \textbf{84} & \underline{78} & \textbf{63} & \textbf{88} & 77 & \textbf{76} & \textbf{79.3}
& \textbf{71} & \textbf{67} & 90 & 69 & \textbf{57} & \underline{53} & 62 & \textbf{67.0} \\
\bottomrule
\end{tabular}%
}
\end{table*}

\subsection{Detailed Results for Reasoning Decomposition}
\label{app:decomposition}
Table~\ref{tab:reasoning_decomposition} reports per-dataset accuracy of single-agent reasoning, where one agent receives all four evidence sources jointly, and multi-agent reasoning with LLM Aggregation, using the same 100 sampled instances per dataset and task.

\begin{table}[t]
\centering
\caption{
Per-dataset accuracy (\%) of single-agent and multi-agent reasoning over the same four evidence sources. Single-agent reasoning receives all evidence sources jointly in one reasoning context; multi-agent reasoning uses LLM Aggregation.  Avg.\ denotes the mean across the seven datasets. The best results within each column are shown in \textbf{bold}.
}
\label{tab:reasoning_decomposition}
\small
\setlength{\tabcolsep}{5pt}
\begin{tabular}{lcccccccc}
\toprule
Method & arXiv & products & PubMed & Reddit & sports & arXiv-23 & computers & Avg. \\
\midrule
\multicolumn{9}{c}{\textit{Link Prediction}} \\
\midrule
Single-agent reasoning & 78 & 83 & 76 & 57 & 75 & \textbf{83} & 70 & 74.6 \\
Multi-agent reasoning  & \textbf{87} & \textbf{85} & \textbf{77} & \textbf{68} & \textbf{89} & 77 & \textbf{77} & \textbf{80.0} \\
\midrule
\multicolumn{9}{c}{\textit{Node Classification}} \\
\midrule
Single-agent reasoning & 65 & 68 & 88 & 64 & 57 & \textbf{54} & 63 & 65.6 \\
Multi-agent reasoning  & \textbf{66} & \textbf{69} & \textbf{91} & \textbf{71} & \textbf{58} & \textbf{54} & \textbf{65} & \textbf{67.7} \\
\bottomrule
\end{tabular}
\end{table}

\subsection{Coordination under Specialist Disagreement}
\label{app:agreement}

Table~\ref{tab:agreement_full} details the performance of each coordination method across levels of initial specialist agreement on both link prediction and node classification. Agreement is measured by the
largest number of specialists producing the same initial prediction. The results show that coordination has little effect under
unanimous agreement, while differences grow as agreement decreases. 

\begin{table}[t]
\centering
\caption{
Accuracy (\%) conditioned on agreement among the four graph specialists.
Percentages in parentheses denote the share of instances in each agreement
group. The best and second-best results within each group are shown in
\textbf{bold} and \underline{underlined} respectively.
}
\label{tab:agreement_full}
\resizebox{\textwidth}{!}{
\begin{tabular}{lcccccc}
\toprule
& \multicolumn{3}{c}{\textbf{Link Prediction}}
& \multicolumn{3}{c}{\textbf{Node Classification}} \\
\cmidrule(lr){2-4}
\cmidrule(lr){5-7}
Method
& $4/4$ (54.5\%)
& $3/4$ (31.4\%)
& $\leq 2/4$ (14.1\%)
& $4/4$ (67.5\%)
& $3/4$ (19.0\%)
& $\leq 2/4$ (13.5\%) \\
\midrule
Majority
& \textbf{88.7}
& 68.9
& 53.9
& \textbf{79.7}
& 49.7
& 39.5 \\
LLM Agg.
& \textbf{88.7}
& \textbf{71.4}
& \underline{66.5}
& \textbf{79.7}
& 49.4
& 39.1 \\
SOP
& 84.3
& 64.6
& 56.6
& 78.7
& 49.5
& 36.1 \\
Debate
& \textbf{88.7}
& 64.3
& 50.1
& \textbf{79.7}
& 47.7
& 38.5 \\
Routing
& \textbf{88.7}
& \underline{71.1}
& \textbf{69.4}
& \textbf{79.7}
& 50.0
& \underline{39.7} \\
DyLAN
& 87.9
& 65.9
& 55.3
& \underline{79.5}
& \textbf{51.5}
& \textbf{41.1} \\
GoA
& \underline{88.6}
& 69.7
& 63.7
& \underline{79.5}
& \underline{51.3}
& 38.9 \\
\bottomrule
\end{tabular}
}
\end{table}

\subsection{Additional Backbone Ablation}
\label{app:backbone_ablation}

We further examine whether the observed multi-agent gains persist under
different LLM backbones. In addition to the main Qwen2.5-32B setting, we rerun all inference calls with
\texttt{Qwen/Qwen2.5-7B-Instruct}~\citep{qwen2024qwen25}, a smaller model from the same family, and
\texttt{zai-org/GLM-4-32B-0414}~\citep{glm2024chatglm}, a similarly sized
model from a different family. Due to computational resource constraints, we use a compact subset of
three datasets, including Reddit, Sports, and Computers, and randomly sample 100
instances per dataset and task.

Table~\ref{tab:backbone_ablation} shows that the multi-agent advantage persists
under both backbones. Within this ablation, \emph{Adaptive Routing} achieves
the\textbf{ highest three-dataset average} for both NC and LP under both backbones.
Compared with the strongest single-agent baseline, its gains are \textbf{6.4 and
7.3 points} with Qwen2.5-7B, and \textbf{8.0 and 12.0 points} with GLM-4-32B on NC and
LP, respectively. These results suggest that the benefits of coordinating
graph specialists are not restricted to the backbone used.

\begin{table*}[h]
\centering
\caption{
Backbone ablation on three datasets using 100 randomly sampled instances per dataset
and task. Results report accuracy (\%). Avg.\ is the mean across the three
datasets. Best and second-best distinct values within each backbone and column
are \textbf{bold} and \underline{underlined} respectively.
}
\label{tab:backbone_ablation}

\setlength{\tabcolsep}{2.2pt}
\renewcommand{\arraystretch}{1.00}

\resizebox{\textwidth}{!}{%
\begin{tabular}{
@{}l|
*{3}{c}>{\columncolor{summaryblue}}c|
*{3}{c}>{\columncolor{summaryblue}}c||
*{3}{c}>{\columncolor{summaryblue}}c|
*{3}{c}>{\columncolor{summaryblue}}c
@{}
}
\toprule

\multirow{3}{*}{\textbf{Method}}
& \multicolumn{8}{c||}{\textbf{Qwen2.5-7B}}
& \multicolumn{8}{c}{\textbf{GLM-4-32B}} \\

\cmidrule(lr){2-9}
\cmidrule(lr){10-17}

& \multicolumn{4}{c|}{\textbf{Link Prediction}}
& \multicolumn{4}{c||}{\textbf{Node Classification}}
& \multicolumn{4}{c|}{\textbf{Link Prediction}}
& \multicolumn{4}{c}{\textbf{Node Classification}} \\

\cmidrule(lr){2-5}
\cmidrule(lr){6-9}
\cmidrule(lr){10-13}
\cmidrule(lr){14-17}

& Reddit & sports & computers & Avg.
& Reddit & sports & computers & Avg.
& Reddit & sports & computers & Avg.
& Reddit & sports & computers & Avg. \\

\midrule

\rowcolor{groupgray}
\multicolumn{17}{c}{\textbf{LLM Reasoning}} \\

Chain-of-Thought
& 52 & 60 & 60 & 57.3
& 44 & 32 & 52 & 42.7
& 49 & 65 & 54 & 56.0
& 63 & 68 & 54 & 61.7 \\

\midrule

\rowcolor{groupgray}
\multicolumn{17}{c}{\textbf{Agentic Graph Learning}} \\

Graph-CoT
& 48 & 48 & 45 & 47.0
& 40 & 26 & 46 & 37.3
& 48 & 51 & 57 & 52.0
& 50 & 59 & 54 & 54.3 \\

Search-o1
& 59 & 71 & 64 & 64.7
& \underline{57} & 45 & 48 & 50.0
& 59 & 71 & 59 & 63.0
& 62 & 69 & 62 & 64.3 \\

GraphSearch-F
& \underline{65} & 70 & 63 & 66.0
& 52 & 48 & 51 & 50.3
& 52 & 56 & 58 & 55.3
& 68 & 65 & 67 & 66.7 \\

\midrule

\rowcolor{groupgray}
\multicolumn{17}{c}{\textbf{General Multi-Agent Methods}} \\

Majority Vote
& 55 & 76 & \textbf{69} & 66.7
& 56 & \underline{52} & \underline{58} & 55.3
& 54 & \textbf{84} & 65 & 67.7
& 68 & 73 & 71 & 70.7 \\

LLM Aggregation
& \textbf{67} & 75 & 56 & 66.0
& 53 & 49 & 55 & 52.3
& 60 & \underline{83} & \underline{67} & \underline{70.0}
& \underline{70} & 74 & \underline{73} & \underline{72.3} \\

SOP Workflow
& 61 & 60 & 53 & 58.0
& 52 & 41 & 57 & 50.0
& \underline{62} & 78 & \underline{67} & 69.0
& 69 & \underline{75} & 64 & 69.3 \\

Multi-Agent Debate
& 54 & 70 & 56 & 60.0
& \underline{57} & 41 & \underline{58} & 52.0
& 54 & 82 & 56 & 64.0
& \underline{70} & 72 & 72 & 71.3 \\

DyLAN
& 53 & 73 & 60 & 62.0
& \underline{57} & \textbf{54} & \underline{58} & \underline{56.3}
& 60 & 73 & 66 & 66.3
& 65 & 70 & 67 & 67.3 \\

Graph of Agents
& 62 & \underline{78} & \underline{66} & \underline{68.7}
& 54 & 51 & \textbf{61} & 55.3
& 54 & 71 & 62 & 62.3
& \textbf{71} & 73 & 69 & 71.0 \\

Adaptive Routing
& \textbf{67} & \textbf{84} & \textbf{69} & \textbf{73.3}
& \textbf{58} & 51 & \textbf{61} & \textbf{56.7}
& \textbf{68} & \underline{83} & \textbf{74} & \textbf{75.0}
& \textbf{71} & \textbf{78} & \textbf{75} & \textbf{74.7} \\

\bottomrule
\end{tabular}%
}
\end{table*}

\subsection{Computational Cost Results}
\label{app:computational-cost}

We randomly sampled 100 evaluation instances and ran all seven
coordination methods on this shared sample. Table~\ref{tab:computational-cost} reports per-question costs,
measured under consistent experimental settings across methods
and datasets.

\textbf{Specialist operations} count specialist rollouts
and standalone revision, message, or peer-judgment attempts.
Each rollout counts as one operation, which may include multiple
model requests.

\textbf{LLM calls} count individual prompts dispatched to the model,
including all specialist and coordination requests; prompts within
a batch are counted separately.

\textbf{Input and output tokens} are summed from the actual token IDs
recorded by the inference engine. These include repeated input
histories, intermediate generations, and discarded responses,
but exclude retrieval encoder tokens.

\textbf{GPU-seconds} measure allocated GPU time: twice the elapsed
wall time of the measured workload, divided by the number of questions.
This includes inference, retrieval, internal queueing, and accounting
overhead, but excludes startup, warmup, and waiting before GPU allocation.
Precomputed corpus embedding costs are excluded.

\begin{table}[h]
\centering
\caption{Per-question computational costs on node classification
(NC) and link prediction (LP). GPU-s denotes allocated A100
GPU-seconds. The minimum
within each task is shown in \textbf{bold}.}
\label{tab:computational-cost}
\small
\setlength{\tabcolsep}{4pt}
\renewcommand{\arraystretch}{1.05}
\begin{tabular}{lrrrrr}
\toprule
Method & Specialist ops & LLM calls & Input tokens & Output tokens & GPU-s \\
\midrule
\multicolumn{6}{c}{\textbf{Node Classification}} \\
\midrule
Adaptive Routing & \textbf{1.80} & \textbf{7.12}
  & \textbf{6,836} & \textbf{787} & \textbf{21.28} \\
Majority Vote      & 4.00  & 8.93  & 9,395  & 1,364 & 33.53 \\
LLM Aggregation    & 4.00  & 9.94  & 10,536 & 1,433 & 36.13 \\
SOP Workflow       & 4.00  & 9.18  & 11,662 & 1,324 & 33.53 \\
Multi-Agent Debate & 5.32  & 11.93 & 13,823 & 1,842 & 42.57 \\
DyLAN              & 4.18  & 9.24  & 9,625  & 1,415 & 34.89 \\
Graph of Agents                & 12.00 & 15.92 & 16,780 & 2,189 & 48.32 \\
\midrule
\multicolumn{6}{c}{\textbf{Link Prediction}} \\
\midrule
Adaptive Routing & \textbf{3.35} & 14.15
  & 15,263 & 1,559 & \textbf{43.73} \\
Majority Vote      & 4.00  & 11.84 & 14,121 & 1,648 & 44.94 \\
LLM Aggregation    & 4.00  & 12.53 & 14,767 & 1,689 & 47.04 \\
SOP Workflow       & 4.00  & \textbf{11.14}
  & 15,395 & \textbf{1,533} & 43.97 \\
Multi-Agent Debate & 6.96  & 19.49 & 26,140 & 2,809 & 74.60 \\
DyLAN              & 4.14  & 11.70 & \textbf{13,666} & 1,649 & 45.25 \\
Graph of Agents                & 12.00 & 17.89 & 19,944 & 2,383 & 58.57 \\
\bottomrule
\end{tabular}
\end{table}
\section{Prompt Templates}
\label{app:prompts}

We provide the core prompt instructions used by GraphMAS for reproducibility.
To improve readability, we omit repeated task inputs, candidate-label lists,
and structured output schemas that are inserted mechanically at runtime.
\textcolor{blue}{Blue text} denotes method-specific content instantiated at
inference time, such as specialist roles or peer responses. Majority Vote
requires no additional LLM prompt and is therefore omitted.

All graph specialists share the same reasoning template and differ only in
their assigned retrieval scope. Coordination methods either append
method-specific instructions to this template or invoke separate
coordination-only LLM calls.

\subsection{Graph Reasoning Specialists}
\label{app:prompt-specialists}

All four specialists share the same reasoning protocol and differ only in
their retrieval scope. Table~\ref{tab:prompt_specialists} summarizes the
shared prompt and role-specific search instructions.

\begin{table*}[h]
\centering
\small
\caption{
Core prompt instructions for GraphMAS graph reasoning specialists.
All specialists share the same reasoning protocol and differ in their
admissible retrieval scope.
}
\setlength{\tabcolsep}{5pt}
\renewcommand{\arraystretch}{1.10}
\begin{tabularx}{\textwidth}{
>{\raggedright\arraybackslash}p{0.20\textwidth}
>{\raggedright\arraybackslash}p{0.15\textwidth}
>{\raggedright\arraybackslash}X}
\toprule
\textbf{Component} & \textbf{Stage} & \textbf{Core Prompt Instruction} \\
\midrule

Graph Specialist
& Reasoning
&
You are the \textcolor{blue}{specialist role}, a reasoning assistant for
graph prediction. You have access to ONE graph search scope:
\textcolor{blue}{assigned scope}. Reason inside
\texttt{<think>...</think>}; use \texttt{<search>...</search>} when
additional graph evidence is needed; after receiving
\texttt{<information>...</information>}, reason over the new evidence before
making the final prediction. \\

\midrule

Proximal Neighborhood
& Retrieval
&
Search the target node's 1-hop and 2-hop neighbors using
\texttt{mode=local} and \texttt{hop=1|2}. \\

\midrule

Distal Neighborhood
& Retrieval
&
Search the target node's 3-hop and 4-hop neighbors using
\texttt{mode=local} and \texttt{hop=3|4}. \\

\midrule

Global Relevance
& Retrieval
&
Search globally structure-relevant nodes selected by Personalized PageRank
using \texttt{mode=global}. \\

\midrule

Semantic Affinity
& Retrieval
&
Search nodes whose attributes are semantically similar to the target node
using \texttt{mode=attribute}. \\

\bottomrule
\end{tabularx}
\label{tab:prompt_specialists}
\end{table*}

\subsection{Static Coordination}
\label{app:prompt-static}

Table~\ref{tab:prompt_static} reports the prompts for LLM Aggregation, SOP
Workflow, and Multi-Agent Debate. LLM Aggregation uses a separate coordinator,
whereas SOP and Debate append peer information to the specialist prompt.

\begin{table*}[h]
\centering
\caption{
Core prompt instructions for static coordination methods.
LLM Aggregation combines independent responses, while SOP and Debate expose
specialists to intermediate peer judgments.
}
\small
\setlength{\tabcolsep}{5pt}
\renewcommand{\arraystretch}{1.10}
\begin{tabularx}{\textwidth}{
>{\raggedright\arraybackslash}p{0.20\textwidth}
>{\raggedright\arraybackslash}p{0.15\textwidth}
>{\raggedright\arraybackslash}X}
\toprule
\textbf{Method} & \textbf{Stage} & \textbf{Core Prompt Instruction} \\
\midrule

LLM Aggregation
& Aggregation
&
You are the coordinator of a multi-agent graph prediction system and have no
search tool. Aggregate the specialist judgments and target information.
Prefer agreement supported by relevant evidence; weight specialists by
confidence and evidence quality.  \\

\midrule

SOP Workflow
& Handoff
&
Earlier specialists reported:
\textcolor{blue}{preceding specialist reports}.
Use their findings as context, then add evidence from YOUR own scope and
correct earlier conclusions when your evidence disagrees. The final
specialist additionally weighs all previous findings and produces the
pipeline's final prediction. \\

\midrule

Multi-Agent Debate
& Revision
&
You previously predicted
\textcolor{blue}{previous prediction}.
Other specialists reported:
\textcolor{blue}{peer responses}.
Re-examine the case from YOUR scope. You may search again within your scope;
revise your answer if persuaded by another specialist, otherwise defend it
with evidence from your scope. \\

\bottomrule
\end{tabularx}

\label{tab:prompt_static}
\end{table*}

\subsection{Adaptive Routing}
\label{app:prompt-routing}

Table~\ref{tab:prompt_routing} shows the routing and final-synthesis prompts.
The router observes previously collected specialist reports, while newly
invoked specialists remain independent of the routing history.

\begin{table*}[h]
\centering
\caption{
Core prompts for Adaptive Routing. Specialists are consulted sequentially,
while the router adapts participation and stopping to the current instance.
}
\small
\setlength{\tabcolsep}{5pt}
\renewcommand{\arraystretch}{1.10}
\begin{tabularx}{\textwidth}{
>{\raggedright\arraybackslash}p{0.20\textwidth}
>{\raggedright\arraybackslash}p{0.15\textwidth}
>{\raggedright\arraybackslash}X}
\toprule
\textbf{Method} & \textbf{Stage} & \textbf{Core Prompt Instruction} \\
\midrule

Adaptive Routing
& Routing
&
You are the adaptive router of a multi-agent graph prediction system.
Available specialists are:
\textcolor{blue}{specialist roles and retrieval scopes}.
After reading the specialist reports collected so far, either CALL another
specialist to gather or corroborate evidence, or ANSWER if the evidence is
already sufficient. At most
\textcolor{blue}{remaining calls}
additional specialist calls are allowed. \\

\midrule

Adaptive Routing
& Final synthesis
&
The routing budget is exhausted. Commit to a final prediction using the
collected specialist reports:
\textcolor{blue}{consulted specialist reports}.
Do not invoke graph retrieval. \\

\bottomrule
\end{tabularx}

\label{tab:prompt_routing}
\end{table*}

\subsection{DyLAN}
\label{app:prompt-dylan}

Table~\ref{tab:prompt_dylan} reports the ranking, revision, and final-selection
prompts used in our DyLAN instantiation. Revision reuses the specialists'
initially retrieved evidence and performs no additional graph retrieval.

\begin{table*}[h]
\centering
\small
\caption{
Core prompts for the DyLAN instantiation. Ranking and revision are triggered
only when the initial specialist responses do not satisfy the early-agreement
criterion.
}
\setlength{\tabcolsep}{5pt}
\renewcommand{\arraystretch}{1.10}
\begin{tabularx}{\textwidth}{
>{\raggedright\arraybackslash}p{0.20\textwidth}
>{\raggedright\arraybackslash}p{0.15\textwidth}
>{\raggedright\arraybackslash}X}
\toprule
\textbf{Method} & \textbf{Stage} & \textbf{Core Prompt Instruction} \\
\midrule

DyLAN
& Ranking
&
Compare the candidate specialist responses according to evidence relevance,
specificity, logical consistency, output validity, robustness to noisy graph
context, and absence of unsupported assumptions. Do not solve the task or
produce a new prediction. Rank the provided responses and retain the
top-$k$ specialists. \\

\midrule

DyLAN
& Revision
&
You are the \textcolor{blue}{retained specialist role}.
Reconsider your previous response using your own specialist evidence and
\textcolor{blue}{other retained responses}. Preserve your assigned role and
do not claim access to graph evidence outside your scope. Incorporate useful
peer evidence and explicitly resolve disagreements. \\

\midrule

DyLAN
& Final selection
&
Compare the retained revised responses and select the single response best
supported by its stated graph evidence. Do not create a new answer or combine
responses. \\

\bottomrule
\end{tabularx}

\label{tab:prompt_dylan}
\end{table*}

\subsection{Graph of Agents}
\label{app:prompt-goa}

Table~\ref{tab:prompt_goa} reports the prompts for peer
scoring, bidirectional message passing, and final pooling. Message-passing
calls reuse the specialists' initially retrieved graph evidence.

\begin{table*}[h]
\centering
\caption{
Core prompts for Graph of Agents, covering peer scoring, bidirectional message passing, and final pooling.
}
\small
\setlength{\tabcolsep}{5pt}
\renewcommand{\arraystretch}{1.08}
\begin{tabularx}{\textwidth}{
>{\raggedright\arraybackslash}p{0.20\textwidth}
>{\raggedright\arraybackslash}p{0.15\textwidth}
>{\raggedright\arraybackslash}X}
\toprule
\textbf{Method} & \textbf{Stage} & \textbf{Core Prompt Instruction} \\
\midrule

Graph of Agents
& Peer scoring
&
As the \textcolor{blue}{judge specialist}, evaluate the other selected
specialists according to evidence relevance, reasoning consistency,
specificity, robustness, and output validity. Assign non-negative relevance
scores that sum to $1$. Do not score yourself or produce a new prediction. \\

\midrule

Graph of Agents
& Source-to-target
&
As the \textcolor{blue}{target specialist}, refine your response using
messages from higher-relevance specialists. Treat peer messages as
communicated conclusions rather than graph evidence you directly retrieved.
Incorporate useful information and reject unsupported claims. \\

\midrule

Graph of Agents
& Target-to-source
&
As the \textcolor{blue}{source specialist}, use the revised responses of
lower-relevance specialists to finalize your own response. Preserve your
original evidence scope and critically incorporate useful corrections or
consensus. \\

\midrule

Graph of Agents
& Pooling
&
Produce the final prediction from
\textcolor{blue}{refined specialist responses and relevance scores}.
Use relevance scores as guidance rather than proof of correctness, and compare
the stated evidence and reasoning. Do not invoke graph tools or introduce new
graph evidence. \\

\bottomrule
\end{tabularx}

\label{tab:prompt_goa}
\end{table*}

\end{document}